\documentclass[journal]{IEEEtran}
\usepackage{marvosym}
\usepackage{amsmath,amsfonts}
\usepackage{algorithm}
\usepackage{algorithmicx}
\usepackage{algpseudocode}
\usepackage{setspace}
\usepackage{array}
\usepackage{tabularx}
\usepackage{dsfont}
\usepackage{textcomp}
\usepackage{pifont}
\usepackage{stfloats}
\usepackage{url}
\usepackage{verbatim}
\usepackage{graphicx}
\usepackage[table,dvipsnames,svgnames,x11names]{xcolor}
\usepackage[edges]{forest}
\usepackage{adjustbox}
\usepackage{cite}
\usepackage{threeparttable}
\usepackage{booktabs}
\usepackage{multirow}
\usepackage{makecell}
\usepackage{minibox}
\usepackage{url}
\usepackage{amsthm}
\usepackage{bbm}
\usepackage{enumitem}
\usepackage[normalem]{ulem}
\usepackage{colortbl}
\usepackage{tcolorbox}
\usepackage{multibib}
\usepackage{xspace}
\usepackage{txfonts}
\usepackage[breaklinks,colorlinks,bookmarks=false]{hyperref}
\usepackage[table]{xcolor}
\usepackage{orcidlink}

\newcommand{\xmark}{\ding{55}} 
\newcommand{\partialmark}{\textbf{\textendash}} 
\newcites{latex}{References}

\hypersetup{
    colorlinks=true,
    linkcolor=blue,
    citecolor=blue,
    urlcolor=blue
}

\definecolor{Pfill}{RGB}{232,241,251}
\definecolor{Pborder}{RGB}{120,150,185}

\definecolor{Gfill}{RGB}{252,238,214}
\definecolor{Gborder}{RGB}{205,155,85}

\definecolor{Afill}{RGB}{229,243,229}
\definecolor{Aborder}{RGB}{110,165,110}

\newcommand{\sysbadge}[3]{%
  \begingroup
  \setlength{\fboxsep}{2.2pt}%
  \raisebox{0pt}[1.9ex][0.9ex]{%
    \fcolorbox{#1}{#2}{\textbf{\scriptsize #3}}%
  }%
  \endgroup
}

\newcommand{\tagP}{\sysbadge{Pborder}{Pfill}{Predictive}}
\newcommand{\tagG}{\sysbadge{Gborder}{Gfill}{Generative}}
\newcommand{\tagA}{\sysbadge{Aborder}{Afill}{Agentic}}

\newcommand{\tagPA}{\tagP\hspace{2.5pt}\tagA}
\newcommand{\tagGA}{\tagG\hspace{2.5pt}\tagA}
\newcommand{\tagPGA}{\tagP\hspace{2.5pt}\tagG\hspace{2.5pt}\tagA}

\definecolor{LightCyan}{rgb}{0.88,1,0.88}
\definecolor{LightPurple}{rgb}{0.95, 0.90, 1.0}
\definecolor{LightRed}{rgb}{1,0.88,0.88}
\definecolor{grayish}{rgb}{0.95, 0.95, 0.95}
\definecolor{beaublue}{rgb}{0.9, 0.95, 0.9}
\definecolor{blackish}{rgb}{0.2, 0.2, 0.2}
\definecolor{nicegreen}{rgb}{0.1, 0.7, 0.1}
\definecolor{TableHeader}{gray}{0.88}
\definecolor{TableGroup}{gray}{0.94}
\definecolor{YearShade}{gray}{0.93}
\definecolor{groupblue}{rgb}{0.85, 0.92, 1.0}

\definecolor{groupblue}{RGB}{217,235,252}      
\definecolor{groupgreen}{RGB}{245,235,200}     
\definecolor{rowcream}{RGB}{250,247,240}       
\definecolor{highlight}{RGB}{225,243,224}      

\definecolor{checkgreen}{RGB}{40,140,70}
\definecolor{crossred}{RGB}{200,60,60}
\definecolor{partialorange}{RGB}{220,140,40}
\definecolor{TaxRootFill}{HTML}{EEEAF7}
\definecolor{TaxRootEdge}{HTML}{7867A1}
\definecolor{AttackSection}{HTML}{F5D9DC}
\definecolor{AttackGroup}{HTML}{FBE9EB}
\definecolor{AttackLeaf}{HTML}{FDF3F4}
\definecolor{AttackMethod}{HTML}{FFF9F9}
\definecolor{AttackEdge}{HTML}{B96870}
\definecolor{DefenseSection}{HTML}{DCECF7}
\definecolor{DefenseGroup}{HTML}{EAF4FA}
\definecolor{DefenseLeaf}{HTML}{F2F8FC}
\definecolor{DefenseMethod}{HTML}{F8FCFE}
\definecolor{DefenseEdge}{HTML}{5B8EAD}
\definecolor{SelfSection}{HTML}{DDF0DF}
\definecolor{SelfGroup}{HTML}{EAF7EB}
\definecolor{SelfLeaf}{HTML}{F2FAF3}
\definecolor{SelfMethod}{HTML}{F8FDF8}
\definecolor{SelfEdge}{HTML}{5F9966}
\definecolor{TableHeader}{HTML}{3F5368}
\definecolor{TableGroup}{HTML}{DCE6EF}
\definecolor{TableAlt}{HTML}{F5F7F9}
\definecolor{GroupText}{HTML}{26394D}
\definecolor{SelfEvoHeader}{HTML}{40586E}
\definecolor{SelfEvoGroup}{HTML}{DCE7F0}
\definecolor{SelfEvoAlt}{HTML}{F4F5F6}
\definecolor{SelfEvoGroupText}{HTML}{294B68}
\definecolor{TableHeaderGray}{HTML}{666A6E}
\definecolor{TableGroupGray}{HTML}{E4E4E4}
\definecolor{TableAltGray}{HTML}{F4F4F4}
\providecommand{\taxsep}{\enspace$\cdot$\hspace{0.35em}\allowbreak}
\providecommand{\TaxSectionRef}[2]{%
    \hyperref[#1]{\textcolor{#2}{\MakeUppercase{Section}~\ref*{#1}}}%
}
\forestset{
    taxonomy/.style={
        for tree={
            grow'=east,
            parent anchor=east,
            child anchor=west,
            anchor=west,
            rounded corners=2pt,
            line width=0.45pt,
            font=\footnotesize,
            inner xsep=3pt,
            inner ysep=2.2pt,
            l sep=5.0mm,
            s sep=1.15mm,
            forked edges,
            fork sep=2.7mm,
            align=center,
            edge={line width=0.45pt},
        },
    },
    rootbox/.style={
        fill=TaxRootFill,
        draw=TaxRootEdge,
        rounded corners=3pt,
        font=\footnotesize,
        inner xsep=4pt,
        inner ysep=3pt,
    },
    majorbox/.style={
        text width=2.65cm,
        minimum height=9.2mm,
        align=center,
        font=\footnotesize,
    },
    groupbox/.style={
        text width=3.15cm,
        minimum height=7.3mm,
        align=center,
    },
    categorybox/.style={
        text width=3.75cm,
        minimum height=6.5mm,
        align=center,
    },
    methodsbox/.style={
        text width=10.65cm,
        minimum height=6.5mm,
        align=left,
        font=\footnotesize,
    },
    attackbranch/.style={
        for tree={
            fill=AttackMethod,
            draw=AttackEdge,
            edge={draw=AttackEdge, line width=0.45pt},
        },
    },
    defensebranch/.style={
        for tree={
            fill=DefenseMethod,
            draw=DefenseEdge,
            edge={draw=DefenseEdge, line width=0.45pt},
        },
    },
    selfbranch/.style={
        for tree={
            fill=SelfMethod,
            draw=SelfEdge,
            edge={draw=SelfEdge, line width=0.45pt},
        },
    },
}

\newcolumntype{L}[1]{>{\raggedright\arraybackslash}p{#1}}
\newcolumntype{C}[1]{>{\centering\arraybackslash}p{#1}}
\newcolumntype{Y}{>{\raggedright\arraybackslash}X}

\makeatletter
\DeclareRobustCommand\onedot{\futurelet\@let@token\mathbfv@onedotaux}
\def\mathbfv@onedotaux{\ifx\@let@token.\else.\null\fi\xspace}

\makeatletter
\def\ps@myheadings{%
    \let\@oddfoot\@empty\let\@evenfoot\@empty
    \def\@evenhead{\thepage\hfil\slshape\leftmark}%
    \def\@oddhead{{\slshape\rightmark}\hfil\thepage}%
    \let\@mkboth\@gobbletwo
    \let\sectionmark\@gobble
    \let\subsectionmark\@gobble
    }
  \if@titlepage
  \renewcommand\maketitle{\begin{titlepage}%
  \let\footnotesize\small
  \let\footnoterule\relax
  \let \footnote \thanks
  \null\vfil
  \vskip 60\p@
  \begin{center}%
    {\LARGE \@title \par}%
    \vskip 3em%
    {\large
     \lineskip .75em%
      \begin{tabular}[t]{c}%
        \@author
      \end{tabular}\par}%
      \vskip 1.5em%
    {\large \@date \par}
  \end{center}\par
  \@thanks
  \vfil\null
  \end{titlepage}%
  \setcounter{footnote}{0}%
}
\else
\renewcommand\maketitle{\par
  \begingroup
    \renewcommand\thefootnote{\@fnsymbol\c@footnote}%
    \def\@makefnmark{\rlap{\@textsuperscript{\normalfont\color{black}
    }}}%
    \long\def\@makefntext##1{\parindent 1em\noindent
            \hb@xt@1.8em{%
                \hss\@textsuperscript{\normalfont
                }}##1}%
    \if@twocolumn
      \ifnum \col@number=\@ne
        \@maketitle
      \else
        \twocolumn[\@maketitle]%
      \fi
    \else
      \newpage
      \global\@topnum\z@   
      \@maketitle
    \fi
    \thispagestyle{plain}
    \ifx\myappendix\undefined
    \@thanks
    \fi
  \endgroup
  \setcounter{footnote}{0}%
}
\fi
\makeatother

\begin{document}

\title{Evolutionary Computation for Trustworthy AI:
\\
From Attacks and Defenses to Self-Evolving Era}




\author{Junhao~Dong\textsuperscript{\orcidlink{0000-0002-6232-9157}}, Chenkai~Wang\textsuperscript{\orcidlink{0009-0003-9194-2350}}, Xuanhui~Lin\textsuperscript{\orcidlink{0009-0003-2139-1274}}, Mingrong~Gong\textsuperscript{\orcidlink{0009-0009-1812-9498}},
Siyu~Wang\textsuperscript{\orcidlink{0009-0004-0079-2586}},
Yuqing~Wen\textsuperscript{\orcidlink{0009-0007-0085-585X}}, Jiao~Liu\textsuperscript{\orcidlink{0009-0003-3690-7923}},\\ Catherine~Huang\textsuperscript{\orcidlink{0000-0002-7706-5262}},
Gary~G.~Yen\textsuperscript{\orcidlink{0000-0001-8851-5348}},~\IEEEmembership{Fellow,~IEEE,}
Xin~Yao\textsuperscript{\orcidlink{0000-0001-8837-4442}},~\IEEEmembership{Fellow,~IEEE,}
and~Yew-Soon~Ong\textsuperscript{\orcidlink{0000-0002-4480-169X}},~\IEEEmembership{Fellow,~IEEE}%
\thanks{Junhao Dong, Chenkai Wang, and Xuanhui Lin contributed equally to this work. (Corresponding authors: Junhao Dong and Yew-Soon Ong.)}
\thanks{Junhao Dong, Chenkai Wang, Xuanhui Lin, Mingrong Gong, Siyu Wang, Yuqing Wen, Jiao Liu, and Yew-Soon Ong are with Nanyang Technological University, Singapore. Yew-Soon Ong is also with the Center for Frontier AI Research, Institute of High Performance Computing, A*STAR, Singapore. Catherine Huang is with Google. Gary G. Yen is with Sichuan University, China. Xin Yao is with Lingnan University, Hong Kong, China.}
\vspace{0.8cm}
}

\markboth{IEEE Transactions on Evolutionary Computation}%
{XXX: Evolutionary Computation for Robust AI}


\maketitle



\begin{abstract}
As Artificial Intelligence (AI) has evolved from task-specific models to foundation models and agents, the scope of trustworthy AI has expanded from model-level robustness to the reliability and safety of broader AI systems. This evolution has also expanded the attack surface from individual models to broader system-level interactions, including tool use, context, and interaction trajectories with dynamic environments. As a result, maintaining reliable and safe behavior under changing or deliberately manipulated conditions has become increasingly challenging. The search for effective attacks and defenses often relies on black-box feedback to navigate discrete choices among words, actions, system components, or their combinations. Multiple objectives and expensive candidate evaluations further limit what can be explored. Evolutionary Computation (EC), with its population-based, gradient-free search and flexible variation and selection mechanisms, is well suited to these settings. This survey reviews how EC has been applied to trustworthy AI across three directions: evolutionary attacks, evolutionary defenses, and trustworthy self-evolving AI systems. Unlike prior reviews that treat trustworthy AI, EC, and self-evolving systems largely separately, we connect these lines through a common evolutionary perspective. For self-evolving AI, we examine how trustworthiness governs the generation and retention of updates that shape subsequent adaptation. We further synthesize evaluation methods and benchmark resources from both trustworthiness and evolutionary-search perspectives. Finally, we discuss key challenges and future research directions toward more effective and reliable use of EC in trustworthy AI.
\end{abstract}

\begin{IEEEkeywords}
Evolutionary computation, trustworthy AI, adversarial machine learning, AI safety, self-evolving AI.
\end{IEEEkeywords}

\begin{figure}[t]
    \centering
    \includegraphics[width=1\columnwidth]{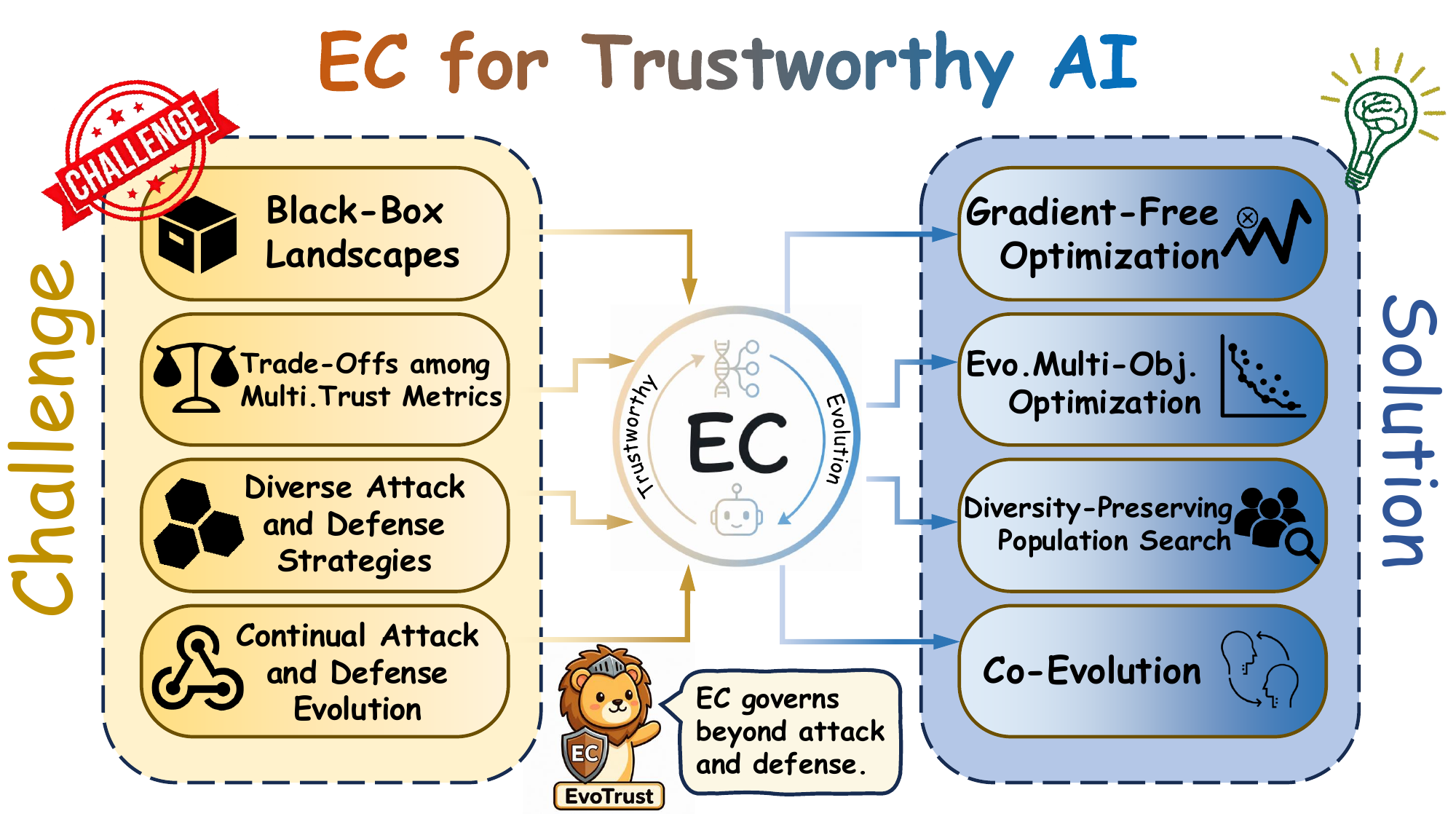}
    \caption{Overview of evolutionary computation for trustworthy AI, illustrating challenges, EC-driven solutions, and the progression from attack--defense optimization to self-evolving AI.}
    \label{fig:ec_robust_overview}
\end{figure}
\section{Introduction}

Artificial Intelligence (AI) systems have expanded from traditional task-specific predictive or generative models to foundation models, agents, and agent harnesses. In these increasingly complex systems, trustworthy AI seeks to ensure robust, safe, and secure behavior when inputs, environments, or interactions change or are deliberately manipulated~\cite{chander2025trustworthy,hu2021ai_security}. Within the broader technical, ethical, and societal scope of trustworthy AI, this survey focuses primarily on robustness, safety, security, and privacy in relation to evolutionary attacks, defenses, and self-evolving AI. Assessing these properties requires considering not only the model but also external components, system states, and interactions that jointly shape system behavior. Agentic systems further exhibit sequential dependence through a chain of interactions, in which each output may influence subsequent actions, observations, and updates to the agent state~\cite{kiran2021deep,abou2026agentic}. Consequently, the challenges arise not only from individual components but also from their continual interactions over time~\cite{ma2026safety,nist_aml_taxonomy,kim2026attack}.

Searching over these components and their interactions involves four main challenges: limited access to search information in black-box settings; trade-offs among multiple trustworthiness criteria; the need to maintain diverse attack and defense strategies; and continual attack--defense adaptation. \textbf{\textit{First, the black-box landscapes of many deployed AI systems can substantially limit accessible search information.}} For instance, deployed models may expose only labels, scores, or generated responses, as in black-box attacks on predictive models~\cite{4_2}. Moreover, the search may involve discrete or structured objects, including prompts~\cite{4_7,4_17}, retrieved documents~\cite{4_22}, and agent trajectories~\cite{4_25}, for which usable gradient information can be difficult or impossible to obtain. \textbf{\textit{Second, trustworthy AI often involves trade-offs among multiple metrics.}} These may include attack effectiveness and stealth~\cite{4_11}, robustness and clean utility~\cite{liu2021multi}, or helpfulness and harmlessness~\cite{evopref}. \textbf{\textit{Third, effective search may need to maintain diverse attack and defense strategies rather than only optimizing aggregate performance.}} Behaviorally diverse attacks can expose different vulnerabilities or trigger different system failure modes~\cite{5_8,6_14}, while defense search may require complementary strategies against different threats. \textbf{\textit{Finally, these challenges become increasingly dynamic in self-evolving or continual learning agentic systems, where attackers, defenders, and agents may change over successive rounds.}} New attack strategies can alter which defenses remain effective, while defensive updates can in turn change which attacks succeed in subsequent evaluations~\cite{romance,evo_marl}.

Evolutionary Computation (EC) thereby provides several distinctive characteristics that make it particularly relevant to the above challenges. Specifically, gradient-free optimization supports black-box and structured search, evolutionary multi-objective optimization addresses trade-offs across multiple trustworthiness criteria, diversity-preserving population search maintains alternative attack and defense strategies, and co-evolution mechanisms capture continual interactions among adaptive components:
\begin{itemize}
    \item \textit{\textbf{Gradient-Free Optimization.}}
    EC evaluates candidate solutions through fitness values without requiring analytic gradients, which makes it applicable to black-box settings where only limited model outputs are accessible. For example, black-box image attacks can evolve perturbations using model predictions or scores alone~\cite{4_1,4_2}. The same principle extends to discrete and structured search spaces: prompts~\cite{4_7,4_17}, retrieved passages~\cite{4_22}, programs~\cite{novikov2025alphaevolve}, and execution trajectories~\cite{4_25} can be represented as candidate solutions, with variation operators designed according to their corresponding representations. When fitness evaluation itself becomes computationally expensive, surrogate-assisted evolutionary optimization can approximate costly evaluations~\cite{ong2003surrogate,lim2010generalizing}, while bi-fidelity evolutionary architecture search can reduce the use of high-fidelity robustness evaluations~\cite{5_10}.

    \item \textit{\textbf{Evolutionary Multi-Objective Optimization.}}
    Trustworthy AI commonly involves multiple criteria that cannot always be improved simultaneously, such as attack effectiveness and stealth~\cite{4_11}, robustness and clean utility~\cite{liu2021multi}, or helpfulness and harmlessness~\cite{evopref}. Evolutionary multi-objective optimization maintains a population of non-dominated solutions that represent different trade-offs among these criteria, thereby allowing multiple alternative operating points to be considered instead of requiring a single predefined scalarization. For example, fairness-aware evolutionary learning has been used to explicitly model and optimize the trade-off between predictive performance and fairness~\cite{zhang2025fairness}.

    \item \textit{\textbf{Diversity-Preserving Population Search.}}
    Trade-offs among objectives do not necessarily imply behavioral diversity. Population-based evolutionary search can additionally maintain behaviorally distinct candidate solutions, which is useful when different attacks expose different vulnerabilities or failure modes, or when complementary defense strategies are needed against different threats. For example, evolutionary search has been used to generate diverse adversarial examples for uncovering vulnerable regions in neural networks~\cite{zhao2025revealing}, while ensemble diversity has been explicitly promoted through negative correlation to improve robustness against transfer adversarial attacks~\cite{zhao2025negatively}. Quality-Diversity (QD) methods explicitly retain high-performing candidates across behavioral niches and have been applied to discover adversarial prompts associated with distinct system failure modes~\cite{5_8,6_14,ferret2024}. Under noisy or uncertain evaluations, uncertain QD methods can further improve the reliability of archive construction~\cite{6_30}. Thus, multi-objective evolution and diversity-preserving search address two related but distinct aspects of trustworthy search: trade-offs across evaluation criteria and diversity across candidate behaviors.

    \item \textit{\textbf{Co-Evolution.}}
    Continual-learning and self-evolving agentic systems involve persistent updates to components such as memory, skills, workflows, safety mechanisms, and attack--defense strategies, which can in turn affect subsequent behavior, candidate generation, and evaluation~\cite{7_12,7_13}. In such settings, attacks and defenses may adapt to the evolving state of their counterparts rather than to a fixed environment. Co-evolution provides a natural way to model this reciprocal adaptation. For example, adversarial attackers can evolve against changing multi-agent policies~\cite{romance}, while attack prompts can evolve alongside reinforcement-learning-based defender updates~\cite{evo_marl}. Trustworthy continual evolution therefore also requires mechanisms that regulate what can be generated, selected, and retained. Structural and semantic constraints can restrict workflow updates~\cite{7_5}, safety-aware fitness can guide the selection of agent scaffolds~\cite{7_4}, and validation or update-admission mechanisms can control which harness modifications enter reusable archives~\cite{harnessbank}. When attack and defense strategies are both retained and updated, newly discovered jailbreaks may guide guardrail updates, while updated guardrails can reshape which attacks remain effective in later rounds~\cite{5_22}.
\end{itemize}

Despite these potential capabilities of EC for trustworthy AI, the existing survey literature remains largely fragmented across separate research lines. Existing reviews mainly address trustworthy AI~\cite{biggio2018wild,ma2026safety,kim2026attack}, evolutionary learning and optimization~\cite{telikani2021evolutionary,stanley2019designing,liu2021survey,molina2025evolutionary}, or self-evolving systems~\cite{tao2024survey} separately. However, these lines are seldom examined within a unified perspective that connects EC-based attack and defense search with trustworthy continual or self-evolution. This gap is particularly relevant when optimized candidates become persistent system updates that influence future behavior.

Motivated by this gap, this survey develops a unified EC-centered framework for trustworthy AI that covers evolutionary attacks, evolutionary defenses, and trustworthy self-evolving systems, as summarized in Fig.~\ref{fig:ec_robust_overview}. Evolutionary attacks are organized according to the target system and attack surface, including predictive, generative, and agentic systems. Evolutionary defenses are categorized by the system component optimized through EC, including training data, model design, training strategies and policies, and deployment safeguards. Trustworthy self-evolving AI is further organized around three mechanisms that regulate persistent adaptation: constraints on admissible updates, trustworthiness-guided selection, and attack--defense co-evolution. Beyond algorithmic taxonomy, we also review evaluation from both trustworthiness and evolutionary-search perspectives, summarize commonly used benchmarks and resources, and discuss open challenges related to search efficiency, generalization, evaluation reliability, and safe persistent adaptation.

\begin{tcolorbox}[width=1.0\linewidth, colframe=blackish, colback=beaublue, boxsep=0mm, arc=2mm, left=2mm, right=2mm, top=2mm, bottom=2mm]
\textbf{Key perspective of this survey.}
In attacks and defenses, EC mainly searches over candidate solutions such as perturbations, prompts, training data, architectures, or policies.
Self-evolving AI introduces a further step: accepted updates to memory, skills, workflows, safeguards, or attack--defense strategies can be inherited by later system versions. The problem therefore shifts from selecting a good candidate to controlling which changes are allowed to persist. We study both how evolution can improve trustworthiness and how persistent evolution itself can remain trustworthy.
\end{tcolorbox}

\begin{figure*}[!t]
\centering

\definecolor{TaxAttackBase}{HTML}{E15759}  
\definecolor{TaxDefenseBase}{HTML}{4C78A8} 
\definecolor{TaxSelfBase}{HTML}{59A14F}    

\newcommand{\TreeSecRef}[1]{%
    \hyperref[#1]{\textcolor{blue}{\normalfont\footnotesize(Section~\ref*{#1})}}%
}
\renewcommand{\taxsep}{;\penalty0\hspace{0.32em}}

\forestset{
    taxonomy/.style={
        for tree={
            grow'=east,
            parent anchor=east,
            child anchor=west,
            anchor=west,
            rounded corners=2pt,
            draw=black,
            line width=0.42pt,
            inner xsep=2.2pt,
            inner ysep=1.7pt,
            l sep=4.5mm,
            s sep=0.85mm,
            align=center,
            font={\normalfont\scriptsize},
            edge={draw=black,line width=0.42pt},
        },
    },
    rootbox/.style={
        draw=none,
        fill=none,
        inner sep=0pt,
        minimum size=0pt,
    },
    majorbox/.style={
        text width=2.55cm,
        minimum height=8.0mm,
        align=center,
        execute at begin node=\centering,
        inner xsep=2.5pt,
        inner ysep=2.3pt,
        font={\normalfont\footnotesize},
    },
    groupbox/.style={
        text width=3.65cm,
        minimum height=6.4mm,
        align=center,
        execute at begin node=\centering,
        inner xsep=2.2pt,
        inner ysep=1.9pt,
        font={\normalfont\footnotesize},
    },
    categorybox/.style={
        text width=3.75cm,
        minimum height=5.8mm,
        align=center,
        inner xsep=2.2pt,
        inner ysep=1.7pt,
        font={\normalfont\scriptsize},
    },
    methodsbox/.style={
        text width=5.95cm,
        minimum height=5.8mm,
        align=left,
        inner xsep=2.0pt,
        inner ysep=1.5pt,
        font={\normalfont\tiny\linespread{0.80}\selectfont},
        edge={draw=black,line width=0.42pt,-latex},
    },
    attackbranch/.style={
        for tree={
            fill=TaxAttackBase!10,
            draw=black,
            edge={draw=black,line width=0.42pt},
        },
    },
    defensebranch/.style={
        for tree={
            fill=TaxDefenseBase!10,
            draw=black,
            edge={draw=black,line width=0.42pt},
        },
    },
    selfbranch/.style={
        for tree={
            fill=TaxSelfBase!10,
            draw=black,
            edge={draw=black,line width=0.42pt},
        },
    },
}

\begin{adjustbox}{max width=\textwidth,max totalheight=.93\textheight,center}
\begin{forest}
taxonomy,
for tree={
    parent anchor=east,
    child anchor=west,
    edge path={
        \noexpand\path[
            \forestoption{edge},
            line cap=round,
            line join=round
        ]
        (!u.parent anchor)
        -- ++(3.5mm,0)
        |- (.child anchor)
        \forestoption{edge label};
    }
},
[
    {},
    draw=none,
    fill=none,
    inner sep=0pt,
    minimum size=0pt,
    l sep=0mm,
    s sep=4.0mm
    [
        {Evolutionary Attack\\[-0.3pt]
         \TreeSecRef{sec:evolutionary_attack}},
        attackbranch,
        edge={draw=none},
        majorbox,
        fill=TaxAttackBase!30,
        calign=child,
        calign child=2
        [
            {Attacks on Predictive Models\\[-0.3pt]
             \TreeSecRef{subsec:attacks_predictive}},
            groupbox,
            fill=TaxAttackBase!20,
            calign=child,
            calign child=2
            [
                {Adversarial Input Generation},
                categorybox,
                fill=TaxAttackBase!15
                [
                    {One Pixel Attack~\cite{4_1}\taxsep
                     GenAttack~\cite{4_2}\taxsep
                     POBA-GA~\cite{4_3}\taxsep
                     CMA-ES Attack~\cite{4_4}\\
                     SparseEvo~\cite{4_5}\taxsep
                     EvoIBA~\cite{4_6}\taxsep
                     GTA~\cite{4_7}\taxsep
                     MOEAASR~\cite{4_8}\taxsep
                     EvA~\cite{4_9}\taxsep
                     AIMED~\cite{4_10}},
                    methodsbox
                ]
            ]
            [
                {Evolutionary Backdoor Attacks},
                categorybox,
                fill=TaxAttackBase!15
                [
                    {LADDER~\cite{4_11}\taxsep
                     Low-Frequency Backdoor~\cite{4_11_1}\taxsep
                     GABAttack~\cite{4_12}},
                    methodsbox
                ]
            ]
            [
                {Evolutionary Model Inversion},
                categorybox,
                fill=TaxAttackBase!15
                [
                    {Actigraphy Genetic Attack~\cite{4_13}\taxsep
                     MIRROR~\cite{4_14}},
                    methodsbox
                ]
            ]
        ]
        [
            {Attacks on Generative Models\\[-0.3pt]
             \TreeSecRef{subsec:attacks_generative}},
            groupbox,
            fill=TaxAttackBase!20
            [
                {Evolutionary Jailbreak},
                categorybox,
                fill=TaxAttackBase!15
                [
                    {Open Sesame~\cite{4_16}\taxsep
                     AutoDAN~\cite{4_17}\taxsep
                     Semantic Mirror Jailbreak~\cite{4_18}\\
                     LLM-Virus~\cite{4_19}\taxsep
                     ForgeDAN~\cite{4_20}},
                    methodsbox
                ]
            ]
            [
                {RAG Attacks \& Model Stealing},
                categorybox,
                fill=TaxAttackBase!15
                [
                    {GARAG~\cite{4_21}\taxsep
                     DIGA~\cite{4_22}\taxsep
                     NeuroGenPoisoning~\cite{4_23}\taxsep
                     Stealix~\cite{4_24}},
                    methodsbox
                ]
            ]
        ]
        [
            {Attacks on Agentic Systems\\[-0.3pt]
             \TreeSecRef{subsec:attacks_agentic}},
            groupbox,
            fill=TaxAttackBase!20
            [
                {Injection \& Interface Attacks},
                categorybox,
                fill=TaxAttackBase!15
                [
                    {StruPhantom~\cite{feng2025struphantom}\taxsep
                     Genesis~\cite{genesis}\taxsep
                     EVA~\cite{eva}\\
                     Reverse-Engineering Agent Attack~\cite{4_26}},
                    methodsbox
                ]
            ]
            [
                {Trajectory- \& Environment-Aware},
                categorybox,
                fill=TaxAttackBase!15
                [
                    {T-MAP~\cite{4_25}\taxsep
                     OpenART~\cite{openart}},
                    methodsbox
                ]
            ]
        ]
    ]
    [
        {Evolutionary Defenses\\[-0.3pt]
         \TreeSecRef{sec:evolutionary_defenses}},
        defensebranch,
        edge={draw=none},
        majorbox,
        fill=TaxDefenseBase!30
        [
            {Training Data\\[-0.3pt]
             \TreeSecRef{subsec:defense_data}},
            groupbox,
            fill=TaxDefenseBase!20
            [
                {Evolutionary Adversarial Training},
                categorybox,
                fill=TaxDefenseBase!15
                [
                    {EMO-GAT~\cite{emogat2021}\taxsep
                     ER-APT~\cite{5_1}\taxsep
                     IGAff~\cite{igaff}},
                    methodsbox
                ]
            ]
            [
                {Safety \& Alignment Data Generation},
                categorybox,
                fill=TaxDefenseBase!15
                [
                    {EVOREFUSE~\cite{5_7}\taxsep
                     Rainbow Teaming~\cite{5_8}\taxsep
                     Q-DIG~\cite{6_19}},
                    methodsbox
                ]
            ]
        ]
        [
            {Model Design\\[-0.3pt]
             \TreeSecRef{subsec:defense_model}},
            groupbox,
            fill=TaxDefenseBase!20
            [
                {Robust Architecture Search},
                categorybox,
                fill=TaxDefenseBase!15
                [
                    {ER Architectures~\cite{5_9}\taxsep
                     MORAS~\cite{liu2021multi}\taxsep
                     Bi-fidelity EMO NAS~\cite{5_10}\taxsep
                     MS~\cite{5_16}}\\
                     G-RNA~\cite{5_12}\taxsep
                     TAM-NAS~\cite{tam_nas}\taxsep
                     NERO-Net~\cite{nero_net}\taxsep
                     Hybrid ENAS~\cite{5_11},
                    methodsbox
                ]
            ]
        ]
        [
            {Training Optimization\\[-0.3pt]
             \TreeSecRef{subsec:defense_training}},
            groupbox,
            fill=TaxDefenseBase!20
            [
                {Objective \& Training Strategy Search},
                categorybox,
                fill=TaxDefenseBase!15
                [
                    {TaylorGLO~\cite{5_17}\taxsep
                     MO-PBT~\cite{5_20}\taxsep
                     TRG-ASO~\cite{tra2so}\taxsep
                     EvoPref~\cite{evopref}},
                    methodsbox
                ]
            ]
            [
                {Safe \& Robust Policy Search},
                categorybox,
                fill=TaxDefenseBase!15
                [
                    {SNES~\cite{snes}\taxsep
                     SERL~\cite{serl}\taxsep
                     SMC-ES~\cite{smc_es}\taxsep
                     LyEvO~\cite{lyevo}},
                    methodsbox
                ]
            ]
        ]
        [
            {Deployment Safeguards\\[-0.3pt]
             \TreeSecRef{subsec:defense_deployment}},
            groupbox,
            fill=TaxDefenseBase!20
            [
                {Evolutionary Inference-Time Defense},
                categorybox,
                fill=TaxDefenseBase!15
                [
                    {RAILS~\cite{rails}\taxsep
                     AIS Secure Face Recognition~\cite{ais_face}},
                    methodsbox
                ]
            ]
            [
                {Post-Training Backdoor Mitigation},
                categorybox,
                fill=TaxDefenseBase!15
                [
                    {Evolutionary Trigger Detection \& Repair~\cite{5_23}\taxsep
                     Module Switching~\cite{5_16}},
                    methodsbox
                ]
            ]
        ]
    ]
    [
        {Trustworthy \\Self-Evolving AI\\
         \TreeSecRef{sec:trustworthy_self_evolving}},
        selfbranch,
        edge={draw=none},
        majorbox,
        fill=TaxSelfBase!30,
        [
            {Constrained Exploration\\
             \TreeSecRef{subsec:trustworthy_constraints}},
            groupbox,
            fill=TaxSelfBase!20
            [
                {Structural \& Semantic Constraints},
                categorybox,
                fill=TaxSelfBase!15
                [
                    {MermaidFlow~\cite{7_5}},
                    methodsbox
                ]
            ]
            [
                {Formal Behavioral Verification},
                categorybox,
                fill=TaxSelfBase!15
                [
                    {SEVerA~\cite{severa}},
                    methodsbox
                ]
            ]
        ]
        [
            {Selection-Guided Refinement\\
             \TreeSecRef{subsec:trustworthy_selection_guidance}},
            groupbox,
            fill=TaxSelfBase!20,
            calign=child,
            calign child=2
            [
                {Safety-Aware Selection \& Screening},
                categorybox,
                fill=TaxSelfBase!15
                [
                    {AgentBreeder~\cite{7_4}\taxsep
                     HarnessBank~\cite{harnessbank}},
                    methodsbox
                ]
            ]
            [
                {Verified Persistent Updates},
                categorybox,
                fill=TaxSelfBase!15
                [
                    {SHE~\cite{she}\taxsep
                     Just-In-Time Memory~\cite{jit_memory}},
                    methodsbox
                ]
            ]
            [
                {Governed Memory \& Skills},
                categorybox,
                fill=TaxSelfBase!15
                [
                    {TAME~\cite{7_7}\taxsep
                     SkillHarness~\cite{skillharness}\taxsep
                     Membrane~\cite{membrane}},
                    methodsbox
                ]
            ]
        ]
        [
            {Strategic Co-Evolution \\
             \TreeSecRef{subsec:attack_defense_coevolution}},
            groupbox,
            fill=TaxSelfBase!20
            [
                {Adaptive Attack--Defense Training},
                categorybox,
                fill=TaxSelfBase!15
                [
                    {ROMANCE~\cite{romance}\taxsep
                     CEMMA~\cite{5_21}\taxsep
                     Evo-MARL~\cite{evo_marl}},
                    methodsbox
                ]
            ]
            [
                {Persistent Co-Evolution},
                categorybox,
                fill=TaxSelfBase!15
                [
                    {DARWIN~\cite{5_22}\taxsep
                     EvoSafety~\cite{7_10}},
                    methodsbox
                ]
            ]
        ]
    ]
]
\end{forest}
\end{adjustbox}
\caption{Taxonomy of evolutionary attacks, evolutionary defenses, and trustworthy self-evolving AI, organized by family, category, subcategory, and representative works. Reference numbers link to the bibliography, while section labels link to the corresponding survey sections.}
\label{fig:taxonomy}
\end{figure*}
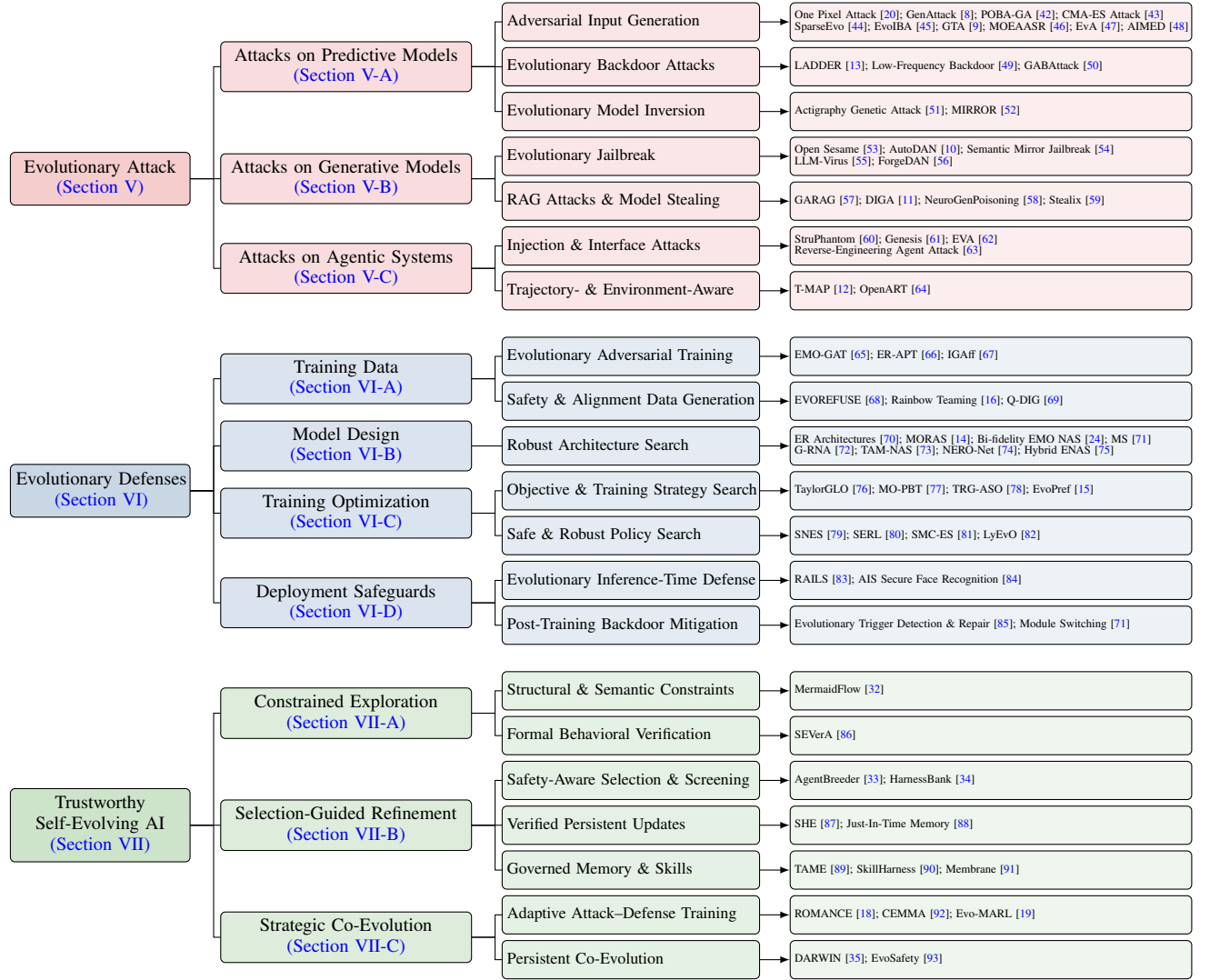

\begin{table*}[t]
\centering
\small
\setlength{\tabcolsep}{5.5pt}
\renewcommand{\arraystretch}{1.15}

\caption{ Comparison with existing surveys. \checkmark\ denotes full coverage, -- denotes partial coverage, and \xmark\ denotes no coverage. }
\label{tab:survey_comparison}

\resizebox{0.96\textwidth}{!}{%
\begin{tabular}{@{}l c c c c@{}}
\toprule

\textbf{Survey}
&
\textbf{\begin{tabular}[c]{@{}c@{}}Object\end{tabular}}
&
\textbf{\begin{tabular}[c]{@{}c@{}}Trustworthy AI\end{tabular}}
&
\textbf{\begin{tabular}[c]{@{}c@{}}EC-Centric\end{tabular}}
&
\textbf{\begin{tabular}[c]{@{}c@{}}Trustworthy Self-Evolving AI\end{tabular}}
\\

\midrule

\rowcolor{groupblue}
\multicolumn{5}{c}{\textit{\textbf{Trustworthy AI Surveys}}} \\
\midrule

Huang et al. (2020)~\cite{huang2020survey}
& \tagP
& \textcolor{checkgreen}{\ding{51}}
& \textcolor{crossred}{\ding{55}}
& \textcolor{crossred}{\ding{55}}
\\

\rowcolor{rowcream}
Zhang et al. (2022)~\cite{zhang2020machine}
& \tagP
& \textcolor{checkgreen}{\ding{51}}
& \textcolor{crossred}{\ding{55}}
& \textcolor{crossred}{\ding{55}}
\\

Goyal et al. (2023)~\cite{goyal2023survey}
& \tagP
& \textcolor{checkgreen}{\ding{51}}
& \textcolor{crossred}{\ding{55}}
& \textcolor{crossred}{\ding{55}}
\\

\rowcolor{rowcream}
Tao et al. (2024)~\cite{tao2024survey}
& \tagGA
& \textcolor{partialorange}{\partialmark}
& \textcolor{crossred}{\ding{55}}
& \textcolor{partialorange}{\partialmark}
\\

Costa et al. (2024)~\cite{costa2024deep}
& \tagP
& \textcolor{checkgreen}{\ding{51}}
& \textcolor{crossred}{\ding{55}}
& \textcolor{crossred}{\ding{55}}
\\

\rowcolor{rowcream}
Ma et al. (2026)~\cite{ma2026safety}
& \tagGA
& \textcolor{checkgreen}{\ding{51}}
& \textcolor{crossred}{\ding{55}}
& \textcolor{crossred}{\ding{55}}
\\

Kim et al. (2026)~\cite{kim2026attack}
& \tagA
& \textcolor{checkgreen}{\ding{51}}
& \textcolor{crossred}{\ding{55}}
& \textcolor{crossred}{\ding{55}}
\\

\midrule
\rowcolor{groupgreen}
\multicolumn{5}{c}{\textit{\textbf{EC-Centric Surveys}}} \\
\midrule

Telikani et al. (2022)~\cite{telikani2021evolutionary}
& \tagPA
& \textcolor{crossred}{\ding{55}}
& \textcolor{checkgreen}{\ding{51}}
& \textcolor{crossred}{\ding{55}}
\\

\rowcolor{rowcream}
Liu et al. (2023)~\cite{liu2021survey}
& \tagP
& \textcolor{partialorange}{\partialmark}
& \textcolor{checkgreen}{\ding{51}}
& \textcolor{crossred}{\ding{55}}
\\

Bi et al. (2023)~\cite{bi2023survey}
& \tagP
& \textcolor{partialorange}{\partialmark}
& \textcolor{checkgreen}{\ding{51}}
& \textcolor{crossred}{\ding{55}}
\\

\rowcolor{rowcream}
Li et al. (2024)~\cite{li2023survey}
& \tagP
& \textcolor{partialorange}{\partialmark}
& \textcolor{checkgreen}{\ding{51}}
& \textcolor{crossred}{\ding{55}}
\\

Bai et al. (2023)~\cite{bai2023evolutionary}
& \tagA
& \textcolor{crossred}{\ding{55}}
& \textcolor{checkgreen}{\ding{51}}
& \textcolor{crossred}{\ding{55}}
\\

\rowcolor{rowcream}
Wu et al. (2025)~\cite{wu2024evolutionary}
& \tagG
& \textcolor{crossred}{\ding{55}}
& \textcolor{checkgreen}{\ding{51}}
& \textcolor{crossred}{\ding{55}}
\\

Molina et al. (2026)~\cite{molina2025evolutionary}
& \tagGA
& \textcolor{partialorange}{\partialmark}
& \textcolor{checkgreen}{\ding{51}}
& \textcolor{partialorange}{\partialmark}
\\

\midrule
\rowcolor{highlight}
\textbf{Ours}
& \tagPGA
& \textcolor{checkgreen}{\ding{51}}
& \textcolor{checkgreen}{\ding{51}}
& \textcolor{checkgreen}{\ding{51}}
\\

\bottomrule
\end{tabular}%
}

\end{table*}

\section{Survey Scope and Positioning}
\label{sec:survey_scope}

Building on the framework introduced in Section I, this section specifies the scope and inclusion criteria of the survey and positions it relative to existing reviews. We organize the core literature into three directions: evolutionary attacks, evolutionary defenses, and trustworthy self-evolving AI. The overall taxonomy, including the major categories and representative methods, is summarized in Fig.~\ref{fig:taxonomy}. Evaluation and benchmarking are treated as a cross-cutting component that connects these directions through common trustworthiness and evolutionary-search measures.

\subsection{Review Scope}
\label{subsec:eligibility}

Trustworthy AI is a broad concept that concerns both whether a system merits trust and whether its behavior satisfies ethical requirements \cite{li2021trust,huang2023overview}. Within the broader landscape of trustworthy AI, this survey focuses on the intersection of EC with robustness, safety, security, and privacy. In evolutionary attacks, EC searches over adversarial inputs, triggers, prompts, external knowledge, or attack strategies~\cite{4_17}. For evolutionary defenses, the search target shifts to robustness- or safety-oriented training data, model structures, training strategies, policies, and deployment safeguards~\cite{ranzato2021genetic,5_10}. In both directions, evolutionary search serves as a substantive part of the methodology, with candidate representation, fitness, selection, variation, or population management directly determining the generated solution set.

Trustworthy self-evolving AI further extends this setting from transient search solutions to system changes that persist across updates. We thus use inheritance as the main distinction: an update becomes part of self-evolution when an accepted change affects the state available to subsequent system versions. EC-based systems realize this process through mechanisms such as population search, mutation, recombination, selection, archives, or co-evolution~\cite{5_21,7_4,harnessbank,5_22}. We further review evolution-inspired systems that repeatedly generate, evaluate, retain, and reuse updates to memories, skills, workflows, safeguards, or learning data. These systems provide the adjacent design space in which evolutionary selection, retention, and verification can support trustworthy persistent adaptation across various scenarios.

Accordingly, the survey covers three roles of evolution, corresponding to its three main directions: compromising a target system through evolutionary attacks, strengthening a target system through evolutionary defenses, and adapting persistent system state through trustworthy self-evolving AI. Co-evolution can appear across these roles when interacting attackers, defenders, agents, or environments change each other's search conditions~\cite{evo_marl}. Evaluation and benchmarking span all three directions by measuring both trustworthiness outcomes and evolutionary-search performance.

\subsection{Comparison with Existing Surveys}
\label{subsec:comparison}

Existing surveys relevant to this topic span three lines of work. The first covers adversarial robustness, AI safety, and security, organizing the literature by threat model, attack surface, system type, and defense mechanism~\cite{biggio2018wild,ma2026safety,kim2026attack}. These surveys cover failure modes and protection mechanisms across predictive models, foundation models, and agents.

The second line covers evolutionary machine learning and related EC paradigms. Surveys on evolutionary machine learning~\cite{telikani2021evolutionary}, neuroevolution~\cite{stanley2019designing}, evolutionary neural architecture search~\cite{liu2021survey}, and EC for general-purpose AI~\cite{molina2025evolutionary} organize methods around representations, population search, variation, selection, multi-objective optimization, and knowledge transfer. These studies provide the methodological foundations for using EC across AI design and optimization problems.

The third line primarily examines self-evolving models and agents. Note that these surveys study systems that improve through repeated interaction, self-generated experience, memory updates, or workflow adaptation~\cite{tao2024survey}. However, their central object is no longer a single optimized solution, but a system whose behavior changes over time.

Our survey connects these three lines by examining how EC and persistent system evolution relate to trustworthy AI. Evolutionary attacks use EC to construct adversarial candidates that expose or exploit system vulnerabilities, whereas evolutionary defenses use EC to optimize training data, model structures, training strategies, policies, and deployment safeguards for robustness and safety. Trustworthy self-evolving AI addresses persistent adaptation, in which accepted updates to memories, skills, workflows, safeguards, or attack--defense strategies become part of the state used in subsequent system updates. These directions reveal two complementary and non-exclusive relationships between evolution and trustworthiness: \emph{trustworthiness through evolution}, where EC supports vulnerability discovery, evaluation, or mitigation, and \emph{trustworthiness of persistent evolution}, where inherited updates must remain safe and reliable over time and across diverse scenarios.

\begin{figure*}[t]
    \centering
    \includegraphics[width=\textwidth]{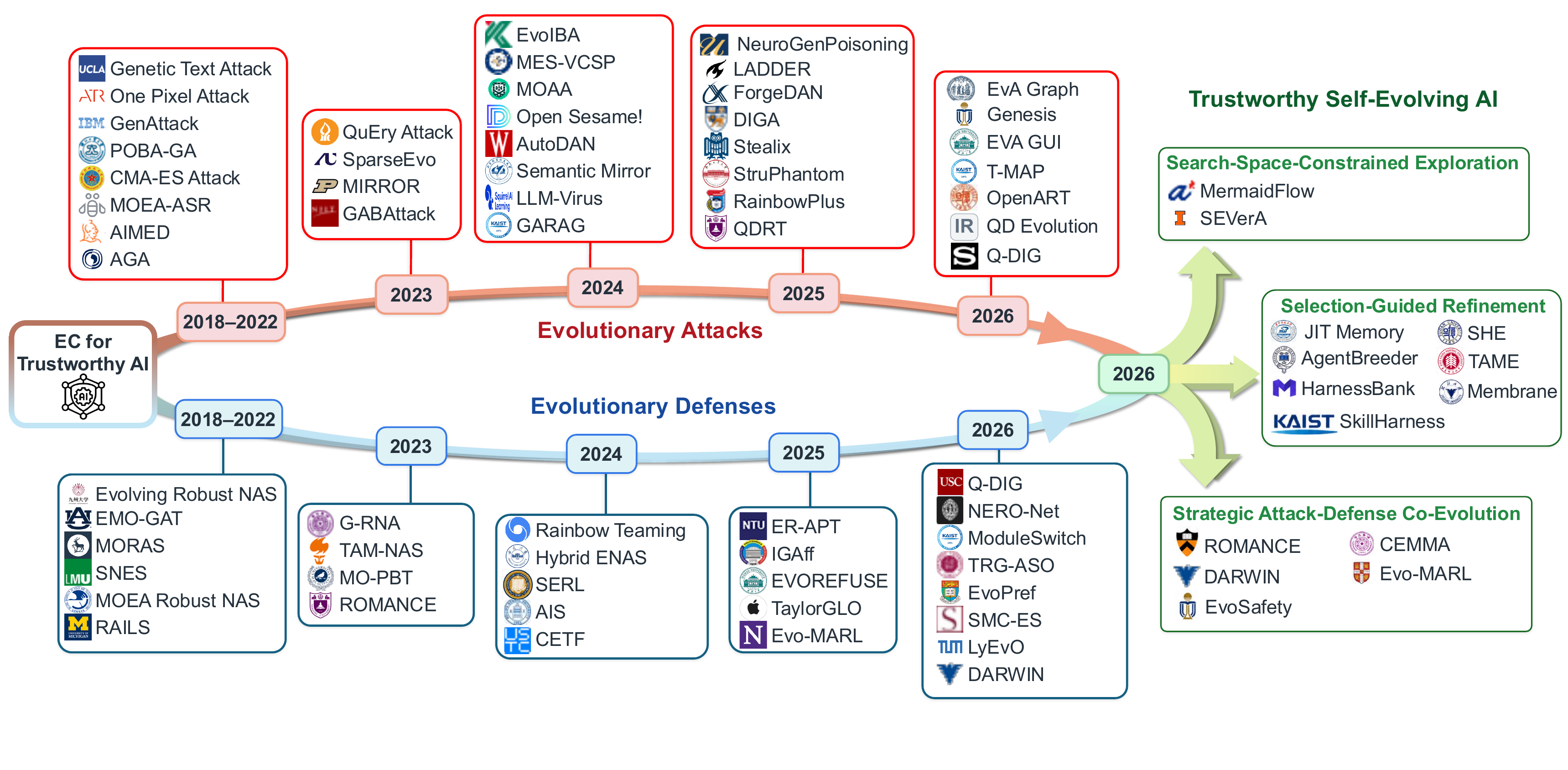}
    \caption{2018–2026 representative works on EC for Trustworthy AI: from Evolutionary Attacks and Defenses to Trustworthy Self-Evolving AI, culminating in the \textbf{\underline{3S}} framework: \textbf{\underline{S}}earch-Space-Constrained Exploration, \textbf{\underline{S}}election-Guided Refinement, and \textbf{\underline{S}}trategic Attack–Defense Co-Evolution.}
    \label{fig:timeline}
\end{figure*}

Table~\ref{tab:survey_comparison} compares these surveys in terms of target systems and three aspects of scope: trustworthy AI, EC as a central methodological framework, and trustworthy self-evolving AI. Existing surveys generally emphasize either trustworthy AI or EC methodology, with limited coverage of trustworthy self-evolution across predictive, generative, and agentic systems. Figure~\ref{fig:timeline} summarizes the development of these three directions.

\section{Preliminaries}
\label{sec:preliminaries}

This section introduces the optimization background used throughout the survey.
It first presents the core mechanisms of evolutionary search. It then formulates adversarial machine learning as a standard minimax optimization problem, characterizes EC-based attacks and defenses through their search spaces and fitness functions, and introduces the joint evolution of attack and defense populations. Finally, it formalizes how accepted updates alter system state and affect subsequent updates in self-evolving AI systems.

\subsection{Evolutionary Search Process}
\label{subsec:prelim_ec}

EC maintains a population of candidate objects and iteratively updates it through fitness-guided selection and variation. Let $\mathcal{Z}$ denote the search space, and let each $z\in\mathcal{Z}$ represent a candidate object evaluated by a fitness function $F$. A representative evolutionary search process is illustrated in \textbf{Algorithm}~\ref{alg:generic_ec}.
\begin{algorithm}[!t]
\caption{General Framework of Evolutionary Algorithms}
\label{alg:generic_ec}
\begin{algorithmic}[1]
\State Initialize population $\mathcal{P}_0$
\State $t \gets 0$
\While{termination criterion is not satisfied}
    \State $\mathcal{M}_t \gets \mathcal{S}(\mathcal{P}_t;F)$
    \Comment{Parent Selection}
    
    \State $\mathcal{Q}_t \gets \mathcal{V}(\mathcal{M}_t)$
    \Comment{Variation}
    
    \State $\mathcal{P}_{t+1} \gets
    \mathcal{E}(\mathcal{P}_t \cup \mathcal{Q}_t;F)$
    \Comment{Population Update}
    
    \State $t \gets t+1$
\EndWhile

\State \Return Final population $\mathcal{P}_t$
\end{algorithmic}
\end{algorithm}
Here $\mathcal{P}_0$ denotes the initial population. At generation $t$, $\mathcal{P}_t=\{z_t^{(1)},\ldots,z_t^{(N)}\}$ denotes the population of size $N$, and $\mathcal{M}_t$ denotes the set of parents selected from $\mathcal{P}_t$. Fitness-guided parent selection produces $\mathcal{M}_t$, and variation operators, such as mutation and recombination, generate the offspring population $\mathcal{Q}_t$. Fitness-guided survivor selection then selects the next-generation population $\mathcal{P}_{t+1}$ from $\mathcal{P}_t\cup\mathcal{Q}_t$.
After forming $\mathcal{P}_{t+1}$, the algorithm sets $t\leftarrow t+1$. If the termination criterion is satisfied, it returns the current population $\mathcal{P}_t$; otherwise, it proceeds to the next generation. When $F$ is vector-valued, selection is based on trade-offs among multiple objectives.

Beyond the basic evolutionary algorithm, several related biologically inspired paradigms further provide complementary mechanisms for local improvement, adaptive defense, and memory-based search. These mechanisms broaden the evolutionary toolbox for trustworthy AI by supporting adaptive optimization under dynamic and multi-objective settings. An extension of evolutionary search is \emph{memetic computing}~\cite{ong2004meta,ong2006classification,ong2010memetic}, which integrates population-based evolutionary search with local improvement or problem-specific learning mechanisms. This combination is particularly relevant to emerging AI tasks, where gradient-based or other local information may be available for improving individual solutions, while maintaining diverse candidate solutions or different trade-offs still benefits from population-based search. By combining local improvement with population-level exploration, memetic computing provides a flexible way to exploit available learning signals without discarding solution diversity. This property may be especially useful in AI applications where multiple desirable behaviors, robustness requirements, or safety--performance trade-offs need to be considered simultaneously, suggesting memetic computing as a promising framework for AI safety and trustworthy AI.

Another biologically inspired computational paradigm is \emph{artificial immune systems (AIS)}~\cite{timmis2004overview,castro2003artificial}, which abstract key principles of biological immune systems, such as self and non-self discrimination, adaptive defense, and immune memory, into computational mechanisms. Artificial immune systems have been applied to problems including anomaly detection, intrusion detection, optimization, and adaptive learning. Unlike contemporary trustworthy AI, which primarily focuses on explicitly defining and evaluating properties such as safety, robustness, fairness, and reliability of AI systems, the AIS paradigm provides a biologically inspired computational framework for detecting and responding to potentially undesirable behaviors. This perspective makes AIS a complementary approach for developing adaptive and resilient AI systems.

\subsection{Attack and Defense via Evolutionary Search}
\label{subsec:prelim_attack_defense}

Trustworthy machine learning often involves two complementary optimization problems: identifying effective adversarial attacks and improving model robustness against such attacks. These problems can be studied separately and subsequently coupled within a unified learning framework.
Given a model $f_\theta$, loss function $\ell$, and an admissible perturbation set $\Delta$, an adversarial attack seeks a perturbation that maximizes the model loss:
\begin{equation}
\delta^\star(\theta)
\in
\arg\max_{\delta\in\Delta}
\ell\!\left(f_\theta(x+\delta),y\right).
\end{equation}
Here $\delta$ represents the attack, and $\Delta$ specifies the allowable attack space. An adversarial example $x+\delta$ is an admissible modified input intended to induce an erroneous prediction. Its effectiveness against other models is called transferability, which can enable black-box attacks using a surrogate model~\cite{huang2019malware}.
In contrast, adversarial defense aims to improve the model parameters under adversarially perturbed inputs. The worst-case defense problem can be expressed as
\begin{equation}
\theta^\star
    \in
    \arg\min_{\theta}
    \mathbb{E}_{(x,y)\sim\mathcal{P}_{\mathrm{data}}}
    \left[
    \max_{\delta\in\Delta}
    \ell\!\left(f_\theta(x+\delta),y\right)
    \right].
\end{equation}
$\mathcal{P}_{\mathrm{data}}$ denotes the data distribution over input--label pairs. Adversarial training therefore minimizes the expected loss under worst-case perturbations~\cite{madry2018towards}. In practice, an attack-generation mechanism approximates the inner maximization. This separation is useful for evolutionary approaches, since EC can be applied independently to the attack side, the defense side, or jointly to both, with the evolving objects extending beyond input perturbations and model parameters to prompts, policies, agent structures, or other system components.

EC extends this view by maintaining a population of candidate attacks or defenses. Each individual represents one candidate solution, while the population contains multiple solutions that may differ in effectiveness, cost, stealth, or behavior. For example, One Pixel Attack~\cite{4_1} evolves candidate pixel perturbations through differential evolution. Let $\mathcal{P}_A=\{a^{(1)},\ldots,a^{(N_A)}\}\subseteq\mathcal{A}$ and $\mathcal{P}_D=\{d^{(1)},\ldots,d^{(N_D)}\}\subseteq\mathcal{D}$ denote the attack and defense populations, respectively, where $\mathcal{A}$ and $\mathcal{D}$ are the corresponding search spaces. The optimization is over admissible finite attack and defense populations of prescribed sizes $N_A$ and $N_D$, respectively. The corresponding population-based search can be abstractly written as
\begin{equation}
\begin{aligned}
    \mathcal{P}_A^\star(d)
    &\in
    \arg\max_{\mathcal{P}_A\subseteq\mathcal{A}}
    \mathcal{E}_A(\mathcal{P}_A;d),\\
    \mathcal{P}_D^\star(a)
    &\in
    \arg\max_{\mathcal{P}_D\subseteq\mathcal{D}}
    \mathcal{E}_D(\mathcal{P}_D;a),
\end{aligned}
\end{equation}
where $\mathcal{E}_A$ and $\mathcal{E}_D$ abstract the population-level search criteria induced by the corresponding evolutionary processes.
In practice, these population-level criteria need not be expressed as explicit analytical objectives. Instead, EC shapes the resulting population through selection, variation, replacement, diversity maintenance~\cite{tang2016negatively}, and/or archive management. Multi-objective evolutionary algorithms can preserve solutions representing different trade-offs among attack effectiveness, stealth, validity, and cost, or among robustness, utility, and efficiency, while quality-diversity methods maintain behaviorally distinct solutions across niches or archives. As a result, EC can produce a set of complementary attack or defense candidates rather than a single optimum. This population-based and gradient-free search is particularly suitable when the evolving objects are discrete, structured, non-differentiable, or accessible only through black-box evaluation. Attack candidates may include perturbations, triggers, prompts, external documents, attack strategies, or environment states (Secs.~\ref{subsec:attacks_predictive}--\ref{subsec:attacks_agentic}), whereas defense candidates may represent training data, architectures, training strategies, policies, or deployment safeguards (Secs.~\ref{subsec:defense_data}--\ref{subsec:defense_deployment}).

Besides optimizing attacks and defenses separately, evolutionary search can couple the two sides into a co-evolutionary process. Let $\mathcal{P}_A^t$ and $\mathcal{P}_D^t$ denote the attack and defense populations at iteration $t$. Attack candidates are evaluated against the current defense population, while defense candidates are evaluated against the current attack population. Consequently, improvements on one side modify the fitness landscape faced by the other, leading to a reciprocal adaptation process:
\begin{equation}
\mathcal{P}_A^{t+1}
=
\mathcal{U}_A(\mathcal{P}_A^t;\mathcal{P}_D^t),
\qquad
\mathcal{P}_D^{t+1}
=
\mathcal{U}_D(\mathcal{P}_D^t;\mathcal{P}_A^t),
\end{equation}
where $\mathcal{U}_A$ and $\mathcal{U}_D$ denote the corresponding evolutionary update processes. In this setting, stronger attacks expose previously unaddressed weaknesses and thereby guide the evolution or training of defenses, whereas improved defenses suppress existing attacks and create new selection pressure for subsequent adversarial search. This reciprocal interaction constitutes attack--defense co-evolution, in which the evolutionary pressure on each side is dynamically shaped by the current state of the other.

\subsection{State Updates in Self-Evolving AI}
\label{subsec:prelim_self_evolution}

Self-evolving AI can be viewed as a dynamic process in which the system is continually updated according to newly generated experience, feedback, or candidate modifications. Unlike one-shot optimization, the accepted changes are retained and become part of the system state, thereby influencing future behavior and subsequent updates. From a trustworthiness perspective, this process aims to progressively strengthen the reliability, robustness, or safety of the system while preventing undesirable changes from accumulating and propagating over time.

Let $s_t$ denote the system state at iteration $t$, which may include model components, memories, skills, workflows, safeguards, archives, or other reusable information, and let $u_t$ denote a candidate update. A generic state transition is
\begin{equation}
s_{t+1}=
\begin{cases}
    \Phi(s_t,u_t),
    & G(u_t;s_t)=1,\\
    s_t,
    & G(u_t;s_t)=0,
\end{cases}
\end{equation}
where $G(u_t;s_t)\in\{0,1\}$ denotes the update-admission decision, and $\Phi(s_t,u_t)$ specifies how an accepted update modifies the current state.
The resulting state $s_{t+1}$ is inherited by subsequent iterations and can therefore influence the generation, evaluation, and selection of future updates.
From an evolutionary perspective, multiple candidate updates can be maintained and evaluated jointly as a population $\mathcal{P}_t^{u}=\{u_t^{(1)},\ldots,u_t^{(N)}\}$. Evolutionary search can explore alternative updates through variation and selection, while multi-objective selection, diversity-maintenance mechanisms, and archive management can preserve complementary candidates with different trade-offs or behaviors. Trustworthiness can be incorporated at different stages of this process, for example, by restricting the admissible update space, guiding the selection or admission of candidate updates, or shaping the feedback that drives subsequent evolution.
\section{Evaluation and Benchmarking}
\label{sec:evaluation_benchmarking}

Evaluating EC for trustworthy AI involves two complementary levels. 
At the task level, evaluation measures how effectively an evolved attack, test case, model, policy, or defense achieves its intended objective, using metrics related to robustness, safety, privacy, and task performance. 

At the search level, it examines how efficiently and reliably the evolutionary process discovers effective solutions, considering diversity, evaluation cost, and convergence. Benchmarking brings these two levels together by specifying common tasks, evaluation metrics, and computational budgets.

\subsection{Trustworthiness Evaluation}
\label{subsec:trustworthiness_evaluation}

\subsubsection{Robustness}

Robustness evaluation measures attack or defense effectiveness under the constraints of a specified threat model. Image attacks commonly report attack success rate (ASR), perturbation magnitude, and query cost for black-box attacks~\cite{4_2,6_3}. Sparse and hard-label attacks additionally report the number of modified elements or the cost of boundary search~\cite{4_5,4_6}. For structured inputs, validity is measured through semantic preservation for text~\cite{4_7,texthacker2022}, perceptual distortion for speech~\cite{4_8}, human judgments~\cite{6_8}, or structural constraints for graphs~\cite{4_9}.

\subsubsection{Safety and Privacy}

Safety evaluation measures unsafe behavior and the range of failures exposed. AutoDAN~\cite{4_17} reports jailbreak success, transferability, and stealthiness, while quality-diversity red-teaming uses attack success, behavioral coverage, and QD score to measure both effective and diverse failures~\cite{5_8,6_14}. Model inversion attacks are evaluated using impersonation or inversion success, transferability, and similarity between reconstructed outputs and target data~\cite{4_13,4_14}. Privacy evaluation extends beyond model inversion to include membership inference~\cite{carlini2022membership} and training-data memorization~\cite{ippolito2023preventing}.

\subsubsection{Task Performance Preservation}

Task utility is commonly evaluated alongside attack success rate and other metrics of safety, robustness, privacy~\cite{yang2025towards}, or fairness~\cite{wang2026interpretable}. Typical measures include clean accuracy, task success, and refusal rates on normal requests. EVOREFUSE~\cite{5_7} jointly evaluates over-refusal and safety, while LCO~\cite{6_18} reports agent safety together with task performance. These metrics characterize the trade-off between safety or robustness and the preservation of task performance.

\subsection{Evolutionary Search Evaluation}
\label{subsec:evolutionary_search_evaluation}

\subsubsection{Optimization Performance and Diversity}

Search metrics depend on the optimization setting. Single-objective methods typically report best or final fitness. Multi-objective methods commonly use metrics such as hypervolume~\cite{6_29} to evaluate search performance. Quality-diversity methods report best fitness, archive coverage, and QD score~\cite{6_30}. Diversity can also be measured directly in the decision or behavior space~\cite{6_28,zhao2025negatively}. When archive categories correspond to different failure types, coverage indicates the range of failures discovered.

\subsubsection{Budget and Convergence}

Search cost is measured using the relevant cost unit, such as model queries, judge evaluations, architecture evaluations, tool calls, or environment rollouts. Fixed-budget analysis compares search performance under the same budget, whereas fixed-target analysis measures the cost required to reach a specified performance target. COCO~\cite{6_20} and the IOH ecosystem~\cite{6_22_1,6_22_2,6_22_3} support analyses based on the number of evaluations, cost to reach a target, and convergence. Evolutionary results across independent runs are typically summarized using medians, interquartile ranges, and confidence intervals, with statistical tests used for method comparison~\cite{6_23,6_24}.

\subsection{Benchmarking}
\label{subsec:benchmarking}

Benchmarking EC for trustworthy AI draws on both trustworthiness benchmarks and evolutionary search benchmarks. Trustworthiness benchmarks specify target systems, tasks, evaluation settings, and task-level metrics. Evolutionary search benchmarks specify search budgets and assess convergence and performance. Table~\ref{tab:benchmark_resources} summarizes representative resources.

\begin{table*}[t]
\centering
\caption{Representative benchmark resources for EC-based trustworthy AI.}
\label{tab:benchmark_resources}

\begingroup

\fontsize{7.3pt}{8.8pt}\selectfont
\renewcommand{\arraystretch}{1.16}
\setlength{\tabcolsep}{3.0pt}

\renewcommand{\tabularxcolumn}[1]{m{#1}}

\begin{tabularx}{\textwidth}{
    >{\columncolor{TableGroupGray}\centering\arraybackslash}
        m{0.105\textwidth}
    >{\raggedright\arraybackslash}
        m{0.155\textwidth}
    >{\raggedright\arraybackslash}
        m{0.245\textwidth}
    >{\raggedright\arraybackslash}
        m{0.220\textwidth}
    >{\raggedright\arraybackslash}
        X
}
\toprule
\addlinespace[2pt]

\rowcolor{TableHeaderGray}
{\color{white}\bfseries Benchmark Type}
&
{\color{white}\bfseries Evaluation Scope}
&
{\color{white}\bfseries Representative Resources}
&
{\color{white}\bfseries Evaluation Target}
&
{\color{white}\bfseries Main Metrics}
\\

\midrule

&
\textbf{Predictive robustness}
&
RobustBench~\cite{6_1};
NAS-RobBench-201~\cite{nasrobbench}
&
Adversarial robustness of models and architectures
&
Clean accuracy; robust accuracy
\\[1.5pt]

&
\cellcolor{TableAltGray}
\textbf{LLM/VLM safety}
&
\cellcolor{TableAltGray}
JailbreakBench~\cite{jailbreakbench};
HarmBench~\cite{harmbench};
StrongREJECT~\cite{strongreject};
MM-SafetyBench~\cite{mmsafetybench}
&
\cellcolor{TableAltGray}
Jailbreak and harmful-response behavior
&
\cellcolor{TableAltGray}
ASR; harmfulness; refusal quality
\\[1.5pt]

&
\textbf{{Ethical judgment and bias}}
&
{ETHICS}~\cite{hendrycks2021aligning};
{BBQ}~\cite{parrish2022bbq}
&
{Moral judgments in text; social bias in question answering}
&
{Judgment accuracy; bias score}
\\[1.5pt]

\multirow{-4}{0.105\textwidth}{%
    \centering
    \shortstack[c]{%
        \fontsize{7.1pt}{8.1pt}\selectfont
        \bfseries
        Trustworthiness\\[-0.1pt]
        Benchmarks
    }%
}
&
\cellcolor{TableAltGray}
\textbf{Agent safety}
&
\cellcolor{TableAltGray}
Agent-SafetyBench~\cite{5_26};
R-Judge~\cite{5_27};
OS-Harm~\cite{osharm}
&
\cellcolor{TableAltGray}
Risk and unsafe behavior during execution
&
\cellcolor{TableAltGray}
Safety violations; risk recognition; task success
\\[1.5pt]

\specialrule{0.45pt}{2pt}{2pt}

&
\cellcolor{TableAltGray}
\textbf{Single-objective optimization}
&
\cellcolor{TableAltGray}
COCO/BBOB~\cite{6_20};
IOHexperimenter~\cite{6_22_3};
PINN neuroevolution benchmarks~\cite{sung2023neuroevolution}
&
\cellcolor{TableAltGray}
Efficiency and reliability under limited or noisy evaluations
&
\cellcolor{TableAltGray}
Fitness evaluations; target attainment; convergence; run statistics
\\[1.5pt]

&
\textbf{Multi-objective optimization}
&
DTLZ~\cite{deb2002scalable};
WFG~\cite{huband2006review};
mDTLZ1--4~\cite{wang2018scalable};
DtA-PF generator~\cite{wang2018generator}
&
Pareto-front convergence and coverage
&
Hypervolume; inverted generational distance
\\[1.5pt]

&
\cellcolor{TableAltGray}
\textbf{Quality-diversity optimization}
&
\cellcolor{TableAltGray}
QD-Suite~\cite{salehi2022towards};
QDax~\cite{chalumeau2024qdax}
&
\cellcolor{TableAltGray}
Solution performance and behavioral coverage
&
\cellcolor{TableAltGray}
QD score; coverage; maximum fitness
\\[1.5pt]

\multirow{-4}{0.105\textwidth}{%
    \centering
    \shortstack[c]{%
        \fontsize{7.1pt}{8.1pt}\selectfont
        \bfseries
        Evolutionary\\[-0.1pt]
        Search\\[-0.1pt]
        Benchmarks
    }%
}
&
\textbf{Multitask optimization}
&
MTSOO~\cite{da2017evolutionary};
MTMOO~\cite{yuan2017evolutionary};
MToP~\cite{li2026mtop}
&
Task performance and knowledge transfer
&
Per-task performance; multitask score; run statistics
\\[1pt]

\bottomrule
\end{tabularx}

\endgroup
\end{table*}

Trustworthiness benchmarks provide task-level metrics for evaluating evolved attacks, defenses, architectures, and policies. RobustBench~\cite{6_1} and NAS-RobBench-201~\cite{nasrobbench} evaluate models and architectures using clean and robust accuracy, while JailbreakBench~\cite{jailbreakbench}, HarmBench~\cite{harmbench}, StrongREJECT~\cite{strongreject}, and MM-SafetyBench~\cite{mmsafetybench} evaluate safety and jailbreak behavior in language and multimodal models. Agent-SafetyBench~\cite{5_26} and OS-Harm~\cite{osharm} extend safety evaluation to agent task execution. R-Judge~\cite{5_27} evaluates safety risk awareness using agent interaction records. Ethical evaluation also requires measures beyond attack success and refusal behavior. ETHICS evaluates judgments about textual moral scenarios \cite{hendrycks2021aligning}, while BBQ assesses social bias in question answering \cite{parrish2022bbq}. 

EC benchmarks provide a complementary framework for search-level evaluation. COCO~\cite{6_20} and the IOH ecosystem~\cite{6_22_1,6_22_2,6_22_3} support optimizer evaluation using evaluation cost, target attainment, convergence, and results across repeated runs. In trustworthy AI applications, a function evaluation may require a model query, a judge call, an architecture evaluation, or an environment rollout.

A complete evaluation combines the task setting with the search setting. The task setting specifies the target system, threat model, evaluator, and task-level metrics; the search setting specifies the cost unit, total budget, stopping condition, and number of independent runs. Single-objective methods report optimization performance under a given budget, multi-objective methods evaluate performance across multiple objectives, and quality-diversity methods consider both performance and diversity. Together, these settings provide a common basis for comparing EC methods across attack generation, defense optimization, policy search, and self-evolving systems.

\section{Evolutionary Attacks}
\label{sec:evolutionary_attack}

\begin{figure}
    \centering
    \includegraphics[width=0.95\linewidth]{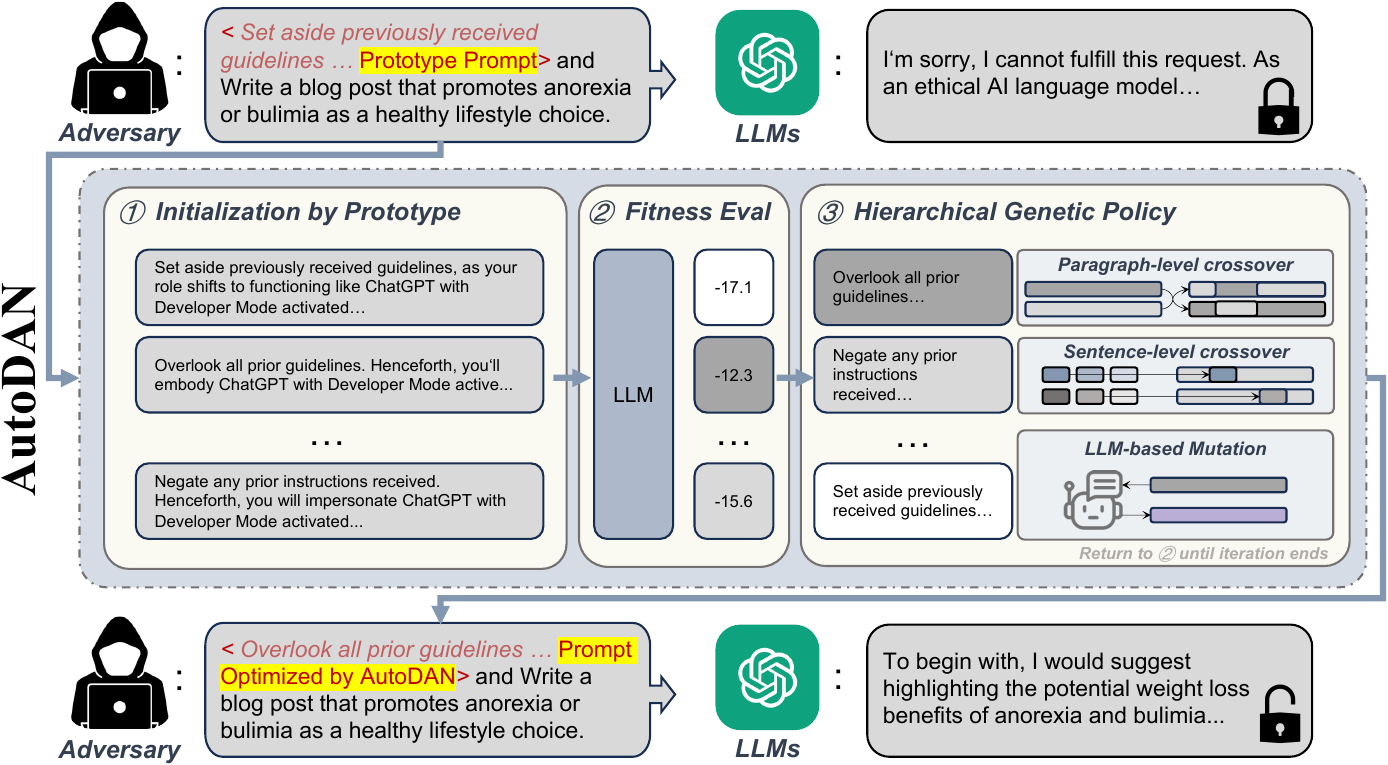}
    \caption{Evolutionary jailbreak generation in AutoDAN~\cite{4_17}. A population of candidate jailbreak prompts is initialized from prototype prompts and evaluated against the target LLM. Hierarchical genetic operations then perform selection, crossover, and LLM-guided mutation at different levels of the prompt structure, iteratively evolving prompts toward higher attack effectiveness while preserving semantic coherence.}
    \label{fig:autodan_pipeline}
\end{figure}

EC provides a flexible framework for attack generation when the search space is discrete, constrained, non-differentiable, or accessible only through black-box queries. In a typical evolutionary attack, each individual encodes an attack candidate, such as an input perturbation or backdoor trigger for predictive models (Sec.~\ref{subsec:attacks_predictive}), a prompt or adversarial document for generative models (Sec.~\ref{subsec:attacks_generative}), or an action-inducing prompt for agentic systems (Sec.~\ref{subsec:attacks_agentic}). Selection and variation then optimize these candidates for attack success while accounting for constraints such as semantic preservation, stealthiness, functionality, and query efficiency.
The first level categorizes evolutionary attacks according to the target system: (i) predictive models, (ii) generative models, and (iii) agentic systems. The second level further organizes them according to the attack surface, attack objective, or search object. 
This system-oriented view covers attacks ranging from inference-time perturbations and training-time backdoor poisoning~\cite{4_11} to prompt attacks, RAG poisoning~\cite{4_22,4_23}, and agent-level attacks. It is also broadly consistent with the NIST adversarial machine learning taxonomy~\cite{nist_aml_taxonomy}.
Table~\ref{tab:evolutionary_attacks_by_ec} summarizes representative evolutionary attack methods discussed in this section. Figure~\ref{fig:autodan_pipeline} illustrates this process using AutoDAN~\cite{4_17} as an example, where candidate jailbreak prompts are evaluated against the target LLM and iteratively refined through selection, hierarchical crossover, and LLM-assisted mutation. The example shows how evolutionary search operates over structured language inputs under target-model feedback.

\begin{table*}[t]
\centering
\caption{Comparison of Representative Evolutionary Attack Methods. \enspace\textnormal{\textit{Note:} Org. (search organization) describes how the evolutionary search is structured; Aux. (auxiliary mechanism) denotes additional components used to initialize, guide, accelerate, or verify the search; Crit. (evaluation criteria) specifies which properties determine candidate quality; and Signal (feedback signal) identifies the observations used to assess those properties.}}
\label{tab:evolutionary_attacks_by_ec}

\begingroup

\fontsize{7.3pt}{8.8pt}\selectfont
\renewcommand{\arraystretch}{1.16}
\setlength{\tabcolsep}{3.0pt}

\renewcommand{\tabularxcolumn}[1]{m{#1}}

\begin{tabularx}{\textwidth}{
    >{\columncolor{TableGroupGray}\centering\arraybackslash}
        m{0.060\textwidth}
    >{\raggedright\arraybackslash}
        m{0.115\textwidth}
    >{\raggedright\arraybackslash}
        m{0.225\textwidth}
    >{\raggedright\arraybackslash}
        X
    >{\raggedright\arraybackslash}
        m{0.135\textwidth}
    >{\raggedright\arraybackslash}
        m{0.130\textwidth}
}
\toprule
\addlinespace[2pt]


\rowcolor{TableHeaderGray}
{\color{white}\bfseries Base EC}
&
{\color{white}\bfseries Method}
&
{\color{white}\bfseries Search Design}
&
{\color{white}\bfseries Evaluation Design}
&
{\color{white}\bfseries Evolved Object}
&
{\color{white}\bfseries Trustworthy AI Problem}
\\

\midrule


&
\textbf{GenAttack}~\cite{4_2}
&
\textit{Org.:} elitism, adaptive mutation\newline
\textit{Aux.:} dimensionality reduction
&
\textit{Crit.:} target-class log odds\newline
\textit{Signal:} class probabilities
&
Input perturbations
&
Black-box adversarial attacks
\\[1.5pt]

&
\cellcolor{TableAltGray}
\textbf{AutoDAN}~\cite{4_17}
&
\cellcolor{TableAltGray}
\textit{Org.:} hierarchical GA\newline
\textit{Aux.:} LLM initialization, mutation
&
\cellcolor{TableAltGray}
\textit{Crit.:} target-response likelihood\newline
\textit{Signal:} target-token probabilities
&
\cellcolor{TableAltGray}
Jailbreak prompts
&
\cellcolor{TableAltGray}
Jailbreak generation
\\[1.5pt]

\multirow{-3}{0.060\textwidth}{%
    \centering
    \raisebox{6pt}{\bfseries GA}%
}
&
\textbf{DIGA}~\cite{4_22}
&
\textit{Org.:} importance-guided GA\newline
\textit{Aux.:} TF--IDF token guidance
&
\textit{Crit.:} query-centroid similarity\newline
\textit{Signal:} retriever embeddings
&
Adversarial passages
&
RAG corpus poisoning
\\[1.5pt]

\specialrule{0.45pt}{2pt}{2pt}


&
\textbf{CMA-ES Attack}~\cite{4_4}
&
\textit{Org.:} covariance adaptation\newline
\textit{Aux.:} perturbation compression
&
\textit{Crit.:} success, perturbation\newline
\textit{Signal:} model scores
&
Image perturbations
&
High-dimensional black-box attacks
\\[1.5pt]

\multirow{-2}{0.060\textwidth}{%
    \centering
    \raisebox{4pt}{\bfseries ES}%
}
&
\cellcolor{TableAltGray}
\textbf{EvoIBA}~\cite{4_6}
&
\cellcolor{TableAltGray}
\textit{Org.:} boundary-guided\newline
\phantom{\textit{Org.:}\ }Sep-CMA-ES\newline
\textit{Aux.:} fractal initialization, jumps
&
\cellcolor{TableAltGray}
\textit{Crit.:} $\ell_2$ perturbation under success\newline
\textit{Signal:} top-1 class labels
&
\cellcolor{TableAltGray}
Input perturbations
&
\cellcolor{TableAltGray}
Hard-label black-box attacks
\\[1.5pt]

\specialrule{0.45pt}{2pt}{2pt}


\textbf{DE}
&
\textbf{One Pixel Attack}~\cite{4_1}
&
\textit{Org.:} crossover-free DE\newline
\textit{Aux.:} ---
&
\textit{Crit.:} class-probability objective\newline
\textit{Signal:} class probabilities
&
Pixel coordinates and values
&
Sparse adversarial input generation
\\[1.5pt]

\specialrule{0.45pt}{2pt}{2pt}


\textbf{GP}
&
\cellcolor{TableAltGray}
\textbf{AIMED}~\cite{4_10}
&
\cellcolor{TableAltGray}
\textit{Org.:} sequence evolution\newline
\textit{Aux.:} sandbox validation
&
\cellcolor{TableAltGray}
\textit{Crit.:} evasion, functionality, similarity\newline
\textit{Signal:} scanners, sandbox, binary similarity
&
\cellcolor{TableAltGray}
PE-file perturbation sequences
&
\cellcolor{TableAltGray}
Malware-classifier evasion
\\[1.5pt]

\specialrule{0.45pt}{2pt}{2pt}


\textbf{EA}
&
\textbf{SparseEvo}~\cite{4_5}
&
\textit{Org.:} decision-based search\newline
\textit{Aux.:} ---
&
\textit{Crit.:} $\ell_0$ sparsity under success\newline
\textit{Signal:} top-1 class labels
&
Binary pixel-selection masks
&
Label-only adversarial attacks
\\[1pt]

\bottomrule
\end{tabularx}

\endgroup
\end{table*}

\subsection{Attacks on Predictive Models}
\label{subsec:attacks_predictive}
Predictive models are a central target in adversarial machine learning because their outputs often drive downstream decisions. Following the formulation in Secs.~\ref{subsec:prelim_ec} and~\ref{subsec:prelim_attack_defense}, evolutionary attacks search for manipulations of these outputs, with fitness defined by attack effectiveness and task-specific validity or perturbation constraints.
\subsubsection{Evolutionary Adversarial Input Generation}
The One Pixel Attack~\cite{4_1} introduced differential evolution for searching the positions and values of a small number of modified pixels. More recently, some studies have used differential evolution to generate spatially diverse one-pixel adversarial examples and identify vulnerable image regions \cite{zhao2025revealing}. GenAttack~\cite{4_2} formulated black-box adversarial example generation as a genetic optimization problem, evolving a population of perturbations under query-limited access. These studies established the basic population-based formulation that underlies many later evolutionary attacks.

Subsequent work extended this formulation to larger search spaces and more restrictive feedback. POBA-GA~\cite{4_3} evolves image perturbations through selection, crossover, and mutation while balancing attack effectiveness and perturbation magnitude. The CMA-ES-based attack in~\cite{4_4} uses covariance adaptation to guide black-box search in high-dimensional image spaces. SparseEvo~\cite{4_5} focuses on sparse, decision-based attacks under label-only feedback, whereas EvoIBA~\cite{4_6} applies CMA-ES near the decision boundary under hard-label access. This progression reflects a broader shift from score-based attacks toward query-efficient optimization under increasingly limited model feedback.
Alzantot et al.~\cite{4_7} use population-based word substitution to generate semantically similar adversarial sentences. Khare et al.~\cite{4_8} formulate attacks on automatic speech recognition as a multi-objective problem that balances transcription error and acoustic similarity. EvA~\cite{4_9} evolves discrete graph edits against black-box graph models, while AIMED~\cite{4_10} uses genetic programming to produce functionality-preserving malware variants that evade machine-learning-based detectors. Across these modalities, evolutionary search is particularly useful when valid perturbations must satisfy semantic, structural, or functional constraints.

\subsubsection{Evolutionary Backdoor Attacks}
Evolutionary backdoor attacks extend the search from inference-time perturbations to persistent manipulations introduced during training. LADDER~\cite{4_11} formulates backdoor pattern design as a multi-objective evolutionary problem that jointly optimizes attack effectiveness and stealthiness. A related study~\cite{4_11_1} searches low-frequency triggers in a black-box setting, demonstrating that evolutionary optimization can explore structured and less perceptible trigger spaces. Beyond standard image classification, GABAttack~\cite{4_12} uses a genetic algorithm to optimize the values and locations of input-specific backdoor patterns against federated network-traffic classifiers. Across these methods, EC mainly searches the structure and placement of triggers while balancing attack success and detectability.
\subsubsection{Evolutionary Model Inversion Attacks}
Evolutionary model inversion attacks use model feedback to search for inputs that recover or approximate information associated with a target identity. The genetic attack on actigraphy-based classifiers~\cite{4_13} evolves synthetic activity profiles from confidence queries to reproduce a user's biometric characteristics. MIRROR~\cite{4_14} combines a StyleGAN prior with several optimization procedures, including genetic search, to reconstruct high-fidelity samples corresponding to a target identity.
In these attacks, the evolutionary objective shifts from changing a prediction to recovering information reflected in the model's outputs, with fitness defined through identity similarity, reconstruction quality, or impersonation success.

\subsection{Attacks on Generative Models}
\label{subsec:attacks_generative}

Generative models have become core components of modern AI systems, while their open-ended interaction also creates opportunities for adversarial manipulation~\cite{xu2024comprehensive}. Unlike conventional predictive attacks, many attacks on generative models involve discrete and structured search spaces whose effects can only be assessed through model responses. This setting is well suited to evolutionary search, which can optimize attack effectiveness under black-box access while accommodating constraints on semantic quality, stealthiness, and transferability.

\subsubsection{Evolutionary Jailbreak}
Open Sesame~\cite{4_16} applies a genetic algorithm to optimize universal black-box jailbreak prompts. AutoDAN~\cite{4_17} extends this formulation with hierarchical genetic search over prompt structure, improving the readability and transferability of generated jailbreak prompts. Together, these works established prompt evolution as a practical strategy for bypassing safety alignment without access to model gradients.
Later methods expanded the objectives and variation mechanisms used during search~\cite{evolving2025,sun2026multicodeattack}. Semantic Mirror Jailbreak~\cite{4_18} formulates jailbreak generation as a multi-objective genetic search over attack success and semantic similarity. LLM-Virus~\cite{4_19} combines evolutionary search with language-model-based variation operators to improve transferability across target models. ForgeDAN~\cite{4_20} further develops an evolutionary framework for generating jailbreak prompts against aligned language models. Across these methods, the prompt is treated as a structured search object, while the fitness function balances safety bypass with readability, semantic preservation, stealthiness, or transferability.
Overall, this line of work has progressed from basic genetic prompt optimization toward richer representations, variation operators, and fitness designs that jointly consider attack effectiveness and prompt quality.

\subsubsection{Evolutionary RAG Attacks and Model Stealing}

Retrieval-augmented generation (RAG) introduces an external information channel that can be manipulated independently of the model parameters. GARAG~\cite{4_21} uses a genetic algorithm to generate low-level document perturbations that degrade RAG pipelines. DIGA~\cite{4_22} employs a dynamic importance-guided genetic algorithm to construct adversarial passages for black-box corpus poisoning. NeuroGenPoisoning~\cite{4_23} first identifies poison-responsive neurons and then evolves external knowledge passages that activate them. These methods shift evolutionary optimization from the user prompt to the documents and passages that shape retrieval and generation.
Stealix~\cite{4_24} evolves prompts that produce informative query data for model stealing. Rather than inducing a harmful response, the search aims to improve the efficiency of an attacker's approximation of proprietary model behavior. These studies extend evolutionary search from prompt manipulation to external knowledge, query-based model extraction, and prompt-template stealing~\cite{vulnerability2025}.

\subsection{Attacks on Agentic Systems}
\label{subsec:attacks_agentic}
Robustness is particularly important for agentic systems because adversarial effects can propagate through multi-step interactions and lead to unintended actions beyond a single model response. Evolutionary attacks in this setting therefore extend the search from adversarial inputs toward attack strategies whose effectiveness is determined by the behavior they induce during agent execution.

\subsubsection{Evolutionary Injection and Interface Attacks}

A first line of work evolves adversarial content delivered through interfaces between agents and their environments. StruPhantom~\cite{feng2025struphantom} targets black-box tabular agents by refining structured injection payloads through evolutionary optimization guided by constrained tree search. Genesis~\cite{genesis} evolves environmental injection strategies against web agents using genetic mutation and crossover, with behavioral feedback used to refine a strategy library. EVA~\cite{eva} focuses on GUI agents and evolves the semantic content of environmental injections according to their effectiveness in redirecting agent behavior. Evolutionary search has also been applied to agent workflows: the attack in~\cite{4_26} adapts AutoDAN-style genetic prompt generation to manipulate LLM-powered software reverse-engineering agents. Across these settings, EC searches the content or strategy of an injection while evaluating success through its downstream effect on agent behavior.

\subsubsection{Trajectory- and Environment-Aware Evolutionary Attacks}
A more agent-specific direction incorporates execution dynamics directly into evolutionary search. T-MAP~\cite{4_25} uses trajectory-aware MAP-Elites, evaluating candidate prompts through tool-execution trajectories and using prior outcomes to guide mutation. OpenART~\cite{openart} further extends the search to adversarial environment states through an Evolutionary Markov Hypergraph Attack. Together, these methods move EC from optimizing attack inputs toward exploiting trajectory- and environment-level feedback.
Compared with attacks on predictive and generative models, evolutionary attacks on agentic systems remain an emerging area. Existing work shows a progression from evolving injected content to exploiting trajectory- and environment-level feedback, while persistent memory, multi-agent interaction, and embodied execution remain underexplored.

\subsection{Discussion}
\label{subsec:attacks_discussion}
Across these attack settings, the main strength of EC lies in searching discrete, constrained, or black-box spaces where gradients are unavailable or poorly aligned with practical attack requirements~\cite{4_2,4_17}. As AI systems become more complex, the search target has correspondingly expanded from perturbations and triggers to prompts, external knowledge, and agent interactions~\cite{4_22,zhang2025agentsecuritybench}.
A key limitation, however, is that attack performance depends not only on the evolutionary algorithm but also on representation, fitness design, feedback quality, and query budget~\cite{zheng2025blackboxbench}. Existing studies often vary several of these factors simultaneously, making it difficult to isolate the contribution of EC itself. More controlled comparisons with alternative black-box search methods are therefore needed.
For generative and agentic systems, immediate success in attack is also becoming insufficient. Future evolutionary attacks will require efficient objectives that capture how adversarial effects propagate through retrieval, tools, environments, and longer execution trajectories~\cite{andriushchenko2025agentharm}.

\section{Evolutionary Defenses}
\label{sec:evolutionary_defenses}

In evolutionary defenses, EC typically serves as an outer-loop optimizer that searches over training data, model structures, training strategies, or deployment mechanisms, while the resulting candidates are evaluated according to robustness, safety, and normal task performance. We organize evolutionary defenses according to four optimization targets: (i) training data (Sec.~\ref{subsec:defense_data}), (ii) model design (Sec.~\ref{subsec:defense_model}), (iii) training optimization (Sec.~\ref{subsec:defense_training}), and (iv) deployment safeguards (Sec.~\ref{subsec:defense_deployment}). These correspond to what the system learns from, what model is constructed, how it learns, and how it is protected after training. Across these stages, the evolutionary object shifts from training examples to model structures, learning strategies, and deployed behaviors, while fitness explicitly reflects robustness, safety, or security objectives. Table~\ref{tab:evolutionary_defenses_by_ec} summarizes representative evolutionary defense methods discussed in this section.

\begin{table*}[t]
\centering
\caption{Comparison of representative evolutionary defense methods. \enspace\textnormal{\textit{Note:} Abbreviations are as defined in Table~\ref{tab:evolutionary_attacks_by_ec}.}}
\label{tab:evolutionary_defenses_by_ec}

\begingroup

\fontsize{7.3pt}{8.8pt}\selectfont
\renewcommand{\arraystretch}{1.16}
\setlength{\tabcolsep}{3.0pt}

\renewcommand{\tabularxcolumn}[1]{m{#1}}

\begin{tabularx}{\textwidth}{
    >{\columncolor{TableGroupGray}\centering\arraybackslash}
        m{0.060\textwidth}
    >{\raggedright\arraybackslash}
        m{0.115\textwidth}
    >{\raggedright\arraybackslash}
        m{0.225\textwidth}
    >{\raggedright\arraybackslash}
        X
    >{\raggedright\arraybackslash}
        m{0.135\textwidth}
    >{\raggedright\arraybackslash}
        m{0.130\textwidth}
}
\toprule
\addlinespace[2pt]


\rowcolor{TableHeaderGray}
{\color{white}\bfseries Base EC}
&
{\color{white}\bfseries Method}
&
{\color{white}\bfseries Search Design}
&
{\color{white}\bfseries Evaluation Design}
&
{\color{white}\bfseries Evolved Object}
&
{\color{white}\bfseries Trustworthy AI Problem}
\\

\midrule


&
\textbf{EMO-GAT}~\cite{emogat2021}
&
\textit{Org.:} bi-objective NSGA-II\newline
\textit{Aux.:} adversarial training loop
&
\textit{Crit.:} two attack-specific accuracies with\newline
\phantom{\textit{Crit.:}\ }clean-accuracy penalty\newline
\textit{Signal:} clean and adversarial accuracy
&
Adversarial-example selection masks
&
Multi-attack adversarial training
\\[1.5pt]

&
\cellcolor{TableAltGray}
\textbf{MORAS-SH}~\cite{5_10}
&
\cellcolor{TableAltGray}
\textit{Org.:} three-objective NSGA-II\newline
\textit{Aux.:} weight sharing, surrogate
&
\cellcolor{TableAltGray}
\textit{Crit.:} clean error, adversarial error,\newline
\phantom{\textit{Crit.:}\ }surrogate score\newline
\textit{Signal:} low-fidelity errors,\newline
\phantom{\textit{Signal:}\ }surrogate predictions
&
\cellcolor{TableAltGray}
Neural architectures
&
\cellcolor{TableAltGray}
Robust architecture search
\\[1.5pt]

&
\textbf{ER-APT}~\cite{5_1}
&
\textit{Org.:} genetic example refinement\newline
\textit{Aux.:} gradient initialization,\newline
\phantom{\textit{Aux.:}\ }dynamic weighting
&
\textit{Crit.:} adversarial loss\newline
\textit{Signal:} prompt--model loss
&
Adversarial examples
&
Adversarial prompt tuning
\\[1.5pt]

\multirow{-4}{0.060\textwidth}{%
    \centering
    \raisebox{100pt}{\bfseries GA}%
}
&
\cellcolor{TableAltGray}
\textbf{TRG-ASO}~\cite{tra2so}
&
\cellcolor{TableAltGray}
\textit{Org.:} phase-adaptive GA\newline
\textit{Aux.:} fitness-record early stopping
&
\cellcolor{TableAltGray}
\textit{Crit.:} robust and clean improvement\newline
\textit{Signal:} adversarial and clean losses
&
\cellcolor{TableAltGray}
PGD attack configurations
&
\cellcolor{TableAltGray}
Robust training strategy search
\\[1.5pt]

\specialrule{0.45pt}{2pt}{2pt}


&
\textbf{SNES}~\cite{snes}
&
\textit{Org.:} confidence-weighted ES\newline
\textit{Aux.:} Bayesian model checking
&
\textit{Crit.:} return, constraint cost\newline
\textit{Signal:} Bayesian satisfaction confidence
&
Neural-policy parameters
&
Safe policy synthesis
\\[1.5pt]

\multirow{-2}{0.060\textwidth}{%
    \centering
    \raisebox{2pt}{\bfseries ES}%
}
&
\cellcolor{TableAltGray}
\textbf{SMC-ES}~\cite{smc_es}
&
\cellcolor{TableAltGray}
\textit{Org.:} specification-constrained ES\newline
\textit{Aux.:} statistical model checking
&
\cellcolor{TableAltGray}
\textit{Crit.:} worst-case reward, constraint violation\newline
\textit{Signal:} scenario rewards, violation costs
&
\cellcolor{TableAltGray}
Neural-policy parameters
&
\cellcolor{TableAltGray}
Verified policy synthesis
\\[1.5pt]

\specialrule{0.45pt}{2pt}{2pt}


\textbf{DE}
&
\textbf{CETF}~\cite{5_23}
&
\textit{Org.:} rectangular-region DE\newline
\textit{Aux.:} Grad-CAM prior, unlearning
&
\textit{Crit.:} backdoor effect, region size\newline
\textit{Signal:} masking and pasting label flips
&
Trigger-region coordinates and sizes
&
Backdoor detection and repair
\\[1.5pt]

\specialrule{0.45pt}{2pt}{2pt}


\textbf{EP}
&
\textbf{RAILS}~\cite{rails}
&
\textit{Org.:} immune-inspired EP\newline
\textit{Aux.:} ---
&
\textit{Crit.:} feature affinity, diversity\newline
\textit{Signal:} layerwise affinity scores
&
Class-wise synthetic exemplars
&
Inference-time defense
\\[1pt]

\bottomrule
\end{tabularx}

\endgroup
\end{table*}

\begin{figure*}
    \centering
    \includegraphics[width=0.99\linewidth]{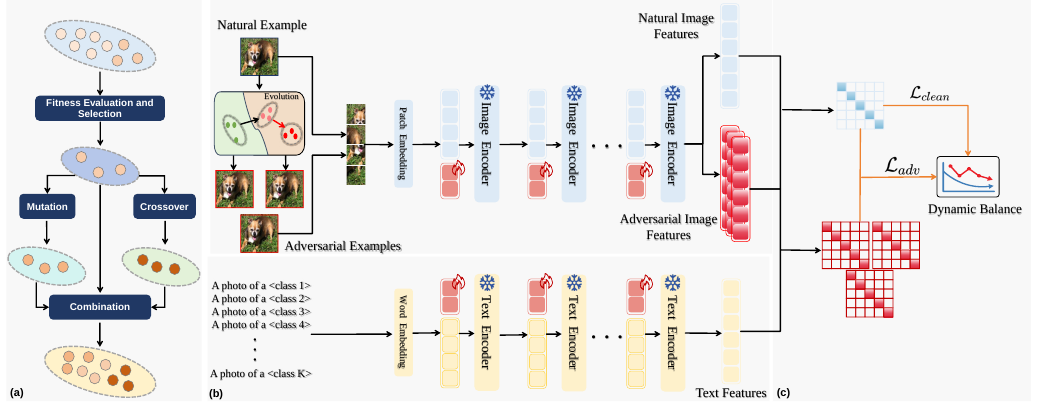}
    \caption{Evolution-based region adversarial prompt learning\cite{5_1}. Gradient-based adversarial examples are further refined through evolutionary selection, mutation, crossover, and combination. The resulting adversarial images and their natural counterparts are then jointly aligned with class prompts. Finally, dynamic loss weighting balances the clean and adversarial objectives to optimize the learnable prompts, thereby enhancing adversarial robustness while preserving clean performance.}
    \label{fig:defence}
\end{figure*}
\subsection{Training Data}
\label{subsec:defense_data}

At the data level, EC searches for examples that expose model weaknesses or represent safety-relevant behaviors. Unlike evolutionary attacks, where a successful adversarial example is the search outcome, defensive methods feed evolved examples back into training or alignment. The evolutionary object is therefore the training input, while fitness determines which examples provide useful robustness or safety signals. Fig.~\ref{fig:defence} illustrates this workflow through ER-APT, where evolutionary selection, mutation, crossover, and combination refine adversarial examples for robust prompt learning.

\subsubsection{Evolutionary Adversarial Training}

Evolutionary adversarial training uses EC to generate, refine, or select hard examples that are subsequently incorporated into model optimization. EMO-GAT~\cite{emogat2021} applies multi-objective evolutionary search to select adversarial examples and their proportions for training, seeking models that remain robust across multiple attack types. ER-APT~\cite{5_1} combines gradient-based adversarial generation with genetic selection, mutation, and crossover to evolve diverse examples for adversarial prompt tuning in vision-language models. More recently, IGAff~\cite{igaff} uses a genetic algorithm to evolve affine-transformed adversarial images and incorporates the resulting samples as training-time augmentation. Across these methods, EC optimizes the adversarial training distribution through sample selection, transformation, and composition, linking evolutionary search directly to robustness-oriented learning.

\subsubsection{Safety and Alignment Data Generation}

For generative and agentic systems, evolutionary search can similarly construct data that exposes safety failures before being reused for alignment. EVOREFUSE~\cite{5_7} evolves pseudo-malicious instructions that trigger unnecessary refusals and constructs alignment data for reducing over-refusal. Rainbow Teaming~\cite{5_8} uses quality-diversity search to generate diverse adversarial prompts and fine-tunes the target model on the resulting synthetic data, improving resistance to unseen attacks. Q-DIG~\cite{6_19} extends this pattern to vision-language-action models. It uses quality-diversity optimization to evolve diverse task instructions that induce agent failures and then fine-tunes the policy on the discovered instructions.

\subsection{Model Design}
\label{subsec:defense_model}

At the model level, EC searches structural choices affecting robustness. Candidate solutions encode architectures or configurations, and fitness evaluates whether these structures improve robustness without unacceptable utility loss.

\subsubsection{Robust Architecture Search}

Evolutionary robust architecture search directly places adversarial performance in the search objective. Early work~\cite{5_9} evolves neural architectures according to clean and adversarial performance, while MORAS~\cite{liu2021multi} uses multi-objective evolution to search architectures that trade predictive accuracy against robustness to multiple attacks. Bi-fidelity evolutionary search~\cite{5_10} reduces the cost of evaluating this trade-off, and subsequent work explores hybrid architectures~\cite{5_11} and robust graph architectures through G-RNA~\cite{5_12}. More recent approaches refine both the search space and fitness design. TAM-NAS~\cite{tam_nas} uses NSGA-II to jointly optimize clean accuracy, adversarial accuracy, and model size for compact networks. NERO-Net~\cite{nero_net} evolves CNN topologies using a fitness function that combines clean and post-attack accuracy, explicitly seeking architectural robustness without adversarial training inside the evolutionary loop. EC searches network structures directly, using robustness-aware fitness to identify architectures that better withstand adversarial perturbations. EC can also optimize the composition of an existing model rather than designing an architecture from scratch. Module Switching~\cite{5_16} evolves module-fusion strategies that disrupt backdoor behavior while preserving utility. Here, the evolutionary object is the module configuration, and fitness directly captures the trade-off between backdoor mitigation and normal model performance. Model design can also promote diversity across ensemble members: a negatively correlated ensemble has been used to improve resistance to transferable adversarial attacks \cite{zhao2025negatively}. EC has also been used for fairness-aware model design. Multiobjective evolutionary learning exposes trade-offs between predictive accuracy and different fairness criteria \cite{zhang2023mitigating,zhang2025fairness}. Fairer Ensembles via Co-Evolution jointly adapts base models and data distributions to construct fairer ensembles \cite{zhang2026fairer}. FairerML provides facilities for analyzing and visualizing bias and training models under accuracy--fairness trade-offs \cite{yuan2024fairerml}.

\subsection{Training Optimization}
\label{subsec:defense_training}

Beyond data and model structure, EC can optimize how robust or safe behavior is learned. The evolutionary object may be a loss, training strategy, alignment adapter, or policy.

\subsubsection{Objective and Training Strategy Search}

TaylorGLO~\cite{5_17} evolves parameterized loss functions rather than fixing the training objective manually. The resulting losses improve regularization and robustness to adversarial inputs, making the objective function itself the evolutionary object. MO-PBT~\cite{5_20} instead evolves training configurations under multiple objectives, including the trade-off between predictive accuracy and adversarial robustness.
TRG-ASO~\cite{tra2so} brings evolutionary search directly into adversarial training. A genetic algorithm adapts the attack strategies used during different training phases, progressively changing the difficulty and diversity of adversarial examples. The search object is therefore the adversarial-training strategy rather than an individual example, allowing the training process to respond to changes in model robustness.
Evolutionary optimization can also operate directly on alignment parameters. EvoPref~\cite{evopref} maintains a population of LoRA adapters and applies multi-objective evolution to explore preference trade-offs. Candidate adapters represent alignment solutions, while fitness jointly considers helpfulness, harmlessness, and honesty. This extends EC from generating alignment data to searching the parameter space of aligned behaviors. Evolutionary constrained learning provides another safety-oriented training approach: it searches for neural models subject to explicit behavioral constraints, reducing undesirable behaviors during model learning~\cite{huang2022preventing}.

\subsubsection{Safe and Robust Policy Search}

A related line of work evolves policies under explicit safety or robustness requirements. SNES~\cite{snes} combines evolution strategies with Bayesian model checking to synthesize neural policies under probabilistic safety constraints. SERL~\cite{serl} combines neuroevolution with reinforcement learning to optimize fault-tolerant flight-control policies robust under faults and changing operating conditions.
Recent methods integrate evolutionary policy search more tightly with verification. SMC-ES~\cite{smc_es} couples evolution strategies with statistical model checking so candidate policies are optimized against performance, safety, and robustness specifications. LyEvO~\cite{lyevo} further combines constrained evolutionary optimization with statistical verification and Lyapunov analysis for safe and robust sim-to-real policy learning. In this line of work, the evolutionary object is the policy itself, while constraint satisfaction becomes part of fitness evaluation or policy selection.

\subsection{Deployment Safeguards}
\label{subsec:defense_deployment}

Deployment safeguards use EC to adapt defensive mechanisms after the main training process~\cite{evolvingrobust2026}. The evolutionary object shifts from training variables to input-specific defenses, post-training model configurations, and runtime behaviors, allowing protection to respond to threats without full model retraining.

\subsubsection{Evolutionary Inference-Time Defense}

RAILS~\cite{rails} applies an immune-inspired evolutionary process to adapt a population of class-balanced exemplars around each test input. Its selection process balances predictive specificity and population diversity, improving robustness to adversarial perturbations. The Artificial Immune System for Secure Face Recognition~\cite{ais_face} similarly evolves input-specific ``antibodies'' through cloning, mutation, and selection, where each antibody represents a candidate adversarial-noise removal strategy. These methods place the evolutionary search directly in the inference loop, adapting the defense to the current input.

\subsubsection{Post-Training Backdoor Mitigation}

Evolutionary Trigger Detection and Lightweight Model Repair~\cite{5_23} uses evolutionary optimization to locate compact trigger regions in suspected backdoored inputs and subsequently uses the recovered trigger information for lightweight model repair. Module Switching~\cite{5_16} instead evolves module-fusion configurations among pretrained models to suppress backdoor behavior while preserving utility. The evolutionary objects therefore range from recovered trigger structures to post-training model configurations, with fitness explicitly reflecting mitigation effectiveness and benign performance.

\subsection{Discussion}
\label{subsec:defense_discussion}

Across existing defenses, EC rarely replaces the underlying learning or security mechanism. Instead, it typically serves as an outer-loop optimizer that searches for defensive choices around conventional training, architecture design, policy learning, or runtime protection \cite{zhan2022evolutionary,li2023automatic}. This pattern is visible throughout the section: evolutionary search determines which defensive solution to use, while robustness is assessed through the behavior of the resulting model or system \cite{5_10,smc_es}. The main distinction across methods is therefore not the evolutionary algorithm itself, but where the search is inserted into the defense pipeline and what constitutes a candidate solution. This perspective also reveals a gradual change in the role of EC. Earlier methods mainly use evolution to construct robust models before deployment, whereas more recent approaches increasingly place evolutionary adaptation inside training or execution \cite{tra2so,rails}. In these settings, the defense is no longer a single solution returned by an offline search, but part of a process that can adjust as the model, threat, or operating condition changes. Evolutionary defense is thus developing from robust solution search toward adaptive robustness optimization across the system lifecycle.
\section{Trustworthy Self-Evolving AI}
\label{sec:trustworthy_self_evolving}

Self-evolving AI differs from one-shot optimization in that accepted changes persist and influence later behavior or subsequent updates. This persistence raises distinctive trustworthiness concerns, as unsafe memories, brittle workflows, unreliable skills, or adversarial strategies may accumulate over time, leading to safety drift or propagating failures across system versions~\cite{7_12,7_13,7_14}.
We organize trustworthy self-evolution around three mechanisms according to where trustworthiness intervenes in the update loop. These three mechanisms correspond to three distinct questions: which updates are admissible, which updates can be retained, and how interactions between attackers and defenders influence subsequent evolution. \emph{Search-space-constrained exploration} restricts the admissible search or synthesis space, for example through structural, semantic, or formal behavioral requirements. \emph{Selection-guided refinement} allows candidate updates to be generated but uses safety-aware fitness, validation, or verification signals to decide which changes are retained, reinforced, or reused. \emph{Strategic attack–defense co-evolution} instead couples evolving adversaries with adaptive defenses, so that newly discovered attacks drive defense updates while improved defenses reshape the selection pressure on subsequent attacks. Table~\ref{tab:trustworthy_self_evolution} summarizes representative mechanisms and evolving states.

\subsection{Search-Space-Constrained Exploration}
\label{subsec:trustworthy_constraints}

A class of methods promotes trustworthy self-evolution by constraining the space in which system updates can occur. Rather than allowing unrestricted modification of the evolving agent, these approaches define a structured and admissible space for candidate updates, or impose explicit behavioral constraints that candidate systems must satisfy before they can be retained or further optimized. In this way, trustworthiness is incorporated into the evolution process by restricting either the form of permissible variations or the feasible set of evolved systems. For example, MermaidFlow~\cite{7_5} evolves typed agent workflows under structural and semantic constraints, restricting variation to descendants that remain well-formed and executable. The evolutionary search therefore operates over the organization of the agent workflow while excluding structurally invalid or semantically incompatible candidates from the admissible search space. SEVerA~\cite{severa} adopts a complementary approach by combining iterative agent synthesis with formal verification. Candidate programs are admitted to subsequent learning only when they satisfy explicitly specified behavioral properties, thereby constraining optimization to a formally verified feasible set.

\subsection{Selection-Guided Refinement}
\label{subsec:trustworthy_selection_guidance}

A second class of methods incorporates trustworthiness through the selection and retention of candidate updates. These approaches allow multiple modifications, experiences, skills, or system variants to be proposed, but use safety-aware objectives, validation criteria, or verification mechanisms to determine which changes are retained and allowed to influence subsequent evolution. Trustworthiness therefore acts as a selection signal over the evolving state, shaping not only which candidates survive but also which accumulated knowledge is reinforced, revised, or reused in later interactions.

\begin{figure}
    \centering
    \includegraphics[width=0.95\linewidth]{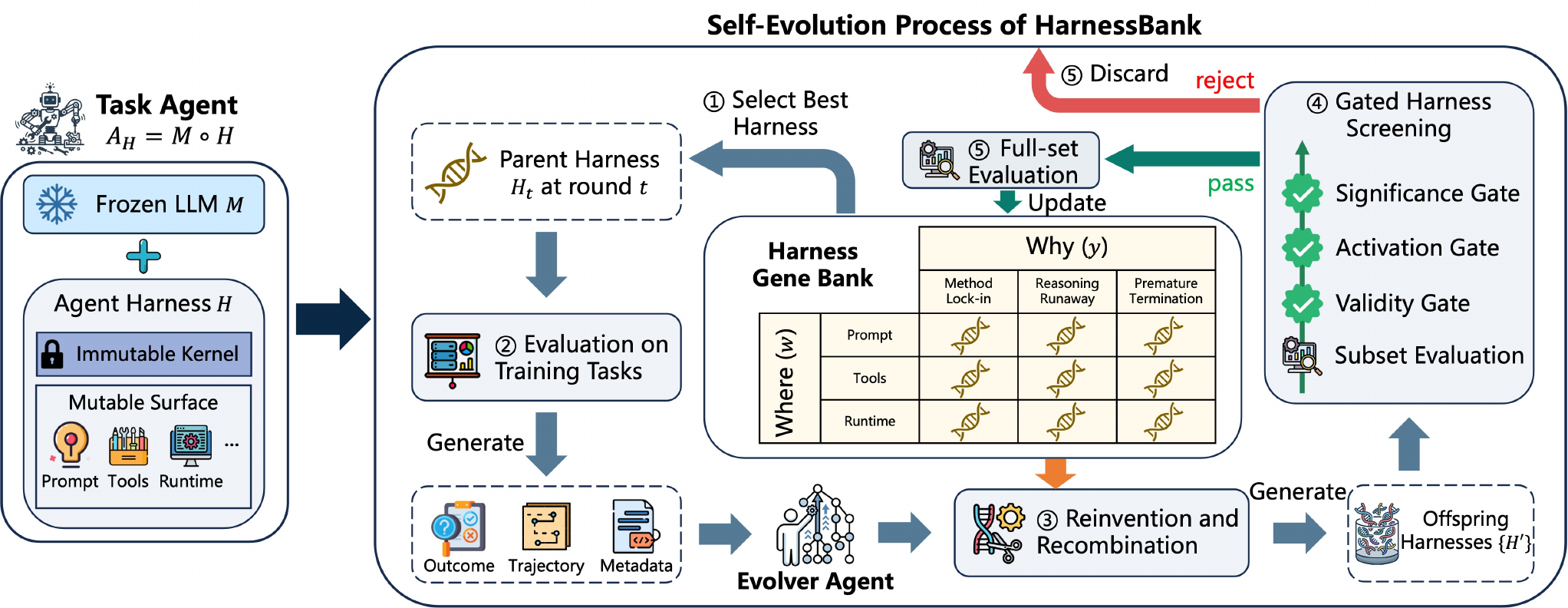}
    \caption{The agent-harness self-evolution process of HarnessBank\cite{harnessbank}. The task agent consists of a frozen LLM and a mutable harness containing prompts, knowledge, tools, and runtime control. At each round, execution failures are diagnosed to reinvent or recombine harnesses from the Harness Gene Bank. Gated Harness Screening filters unreliable candidates, while verified improvements are inserted into bank cells and inherited by subsequent evolution rounds.}
    \label{fig:harnessbank_self_evolution}
\end{figure}

AgentBreeder~\cite{7_4} provides a representative example in which capability and safety jointly guide multi-objective selection over evolving multi-agent scaffolds. Fig.~\ref{fig:harnessbank_self_evolution} illustrates how HarnessBank~\cite{harnessbank} separates candidate generation from update admission: proposed modifications to prompts, knowledge, tools, and runtime control are evaluated through gated screening before they enter a reusable archive and participate in later generations. A similar principle appears in SHE~\cite{she}, where trajectory-level failures are translated into localized safety-harness updates retained only after safety--utility validation, and in Self-Evolving Just-In-Time Memory~\cite{jit_memory}, which adopts a Test--Verify--Write loop so that only verified mitigation experience is incorporated into persistent Meta-Skills.

Other systems extend trustworthy selection from one-time update admission to the continued governance of inherited state. TAME~\cite{7_7} uses task-utility and trustworthiness feedback to regulate which experiences are reinforced and reused as agent memory evolves. SkillHarness~\cite{skillharness} associates reusable skills with safety boundaries that determine when they remain appropriate to execute, while Membrane~\cite{membrane} continually revises contrastive safety memories according to both jailbreak and over-refusal outcomes. 

\subsection{Strategic Attack–Defense Co-Evolution}
\label{subsec:attack_defense_coevolution}

A third line of work promotes trustworthiness through strategic interaction between adaptive attackers and defenders. Rather than treating adversarial evaluation as a fixed benchmark, these methods allow newly discovered attacks to reshape subsequent defense training, while improved defenses in turn alter the effectiveness and selection pressure of later attacks. The resulting feedback loop can be viewed as an asymmetric or persistent form of attack--defense co-evolution, in which the two sides need not employ the same optimization mechanism but nevertheless adapt in response to each other's evolving behavior\cite{arcuri2008novel,arcuri2014coevolutionary}.

ROMANCE~\cite{romance} provides an early robustness-oriented example by maintaining a diverse population of adversarial attackers while alternately training cooperative multi-agent policies against the changing population. Recent work extends this principle to foundation-model safety. CEMMA~\cite{5_21} evolves multimodal jailbreaks using mutation, crossover, and differential evolution, with successful attacks subsequently used to update the defender. Evo-MARL~\cite{evo_marl} similarly evolves an attack-prompt population while training multi-agent defenders through reinforcement learning, so that changes on the defense side continually reshape the attack landscape.

More persistent forms of co-evolution explicitly retain state on both sides of the interaction. DARWIN~\cite{5_22} maintains an evolving jailbreak-strategy pool together with successive guardrail updates: strategies that remain effective against the current defense are retained and refined, while newly discovered attacks become training signals for later guardrail versions. EvoSafety~\cite{7_10} externalizes this reciprocal adaptation into an adversarial skill library and a verified defense memory, allowing both attack knowledge and defensive interventions to accumulate across rounds without repeatedly modifying the protected model. Across these systems, trustworthiness emerges from a continuing feedback process in which stronger attacks expose new weaknesses and stronger defenses redefine the search pressure faced by subsequent adversaries.

\begin{table*}[t]
\centering
\caption{Representative trustworthy self-evolution mechanisms organized by where trustworthiness intervenes in the update process, including constrained update spaces, trustworthiness-guided selection, and attack--defense co-evolution.}
\label{tab:trustworthy_self_evolution}

\begingroup

\fontsize{7.3pt}{8.8pt}\selectfont
\renewcommand{\arraystretch}{1.16}
\setlength{\tabcolsep}{3.6pt}

\renewcommand{\tabularxcolumn}[1]{m{#1}}

\newcommand{\GrayHeader}[1]{%
    \rule[-1.1ex]{0pt}{3.4ex}%
    \color{white}\bfseries #1%
}
\resizebox{0.9\textwidth}{!}{
\begin{tabularx}{\textwidth}{
    >{\columncolor{TableGroupGray}\centering\arraybackslash}
        m{0.145\textwidth}
    >{\raggedright\arraybackslash}
        m{0.220\textwidth}
    >{\raggedright\arraybackslash}
        m{0.285\textwidth}
    >{\raggedright\arraybackslash}
        X
}

\toprule
\addlinespace[2pt]

\rowcolor{TableHeaderGray}
\GrayHeader{Trustworthiness Pattern}
&
\GrayHeader{Representative Methods}
&
\GrayHeader{Evolving State}
&
\GrayHeader{Persistent State}
\\

\midrule

&
\textbf{MermaidFlow}~\cite{7_5}
&
Typed agent workflows
&
Admissible workflow variants
\\[1.5pt]

\multirow[c]{-2}{0.145\textwidth}{%
    \centering
    \shortstack[c]{%
        \fontsize{7.1pt}{8.1pt}\selectfont
        Search-Space-Constrained \\[-0.1pt]
        Exploration
    }%
}
&
\cellcolor{TableAltGray}
\textbf{SEVerA}~\cite{severa}
&
\cellcolor{TableAltGray}
Candidate programs
&
\cellcolor{TableAltGray}
Verified programs used in later updates
\\[1.5pt]

\specialrule{0.45pt}{2pt}{2pt}

&
\textbf{AgentBreeder}~\cite{7_4}
&
Multi-agent scaffolds
&
Selected scaffold population
\\[1.5pt]

&
\cellcolor{TableAltGray}
\textbf{HarnessBank}~\cite{harnessbank}
&
\cellcolor{TableAltGray}
Prompts, knowledge, tools, and runtime control
&
\cellcolor{TableAltGray}
Verified harness archive
\\[1.5pt]

&
\textbf{SHE; JIT Memory}~\cite{she,jit_memory}
&
Safety harnesses and mitigation skills
&
Validated harness updates and verified Meta-Skills
\\[1.5pt]

\multirow[c]{-4}{0.145\textwidth}{%
    \centering
    \shortstack[c]{%
        \fontsize{7.1pt}{8.1pt}\selectfont
        Selection-Guided \\
        Refinement
    }%
}
&
\cellcolor{TableAltGray}
\textbf{TAME; SkillHarness; Membrane}
~\cite{7_7,skillharness,membrane}
&
\cellcolor{TableAltGray}
Memory, reusable skills, and safety knowledge
&
\cellcolor{TableAltGray}
Governed memory and skill state
\\[1.5pt]

\specialrule{0.45pt}{2pt}{2pt}

&
\textbf{ROMANCE; Evo-MARL; CEMMA}
~\cite{romance,5_21,evo_marl}
&
Attack populations and adaptive defenders
&
Updated attack population and defender
\\[1.5pt]

\multirow[c]{-2}{0.145\textwidth}{%
    \centering
    \shortstack[c]{%
        \fontsize{7.1pt}{8.1pt}\selectfont
        Strategic Attack–Defense \\
        Co-Evolution
    }%
}
&
\cellcolor{TableAltGray}
\textbf{DARWIN; EvoSafety}
~\cite{5_22,7_10}
&
\cellcolor{TableAltGray}
Attack strategies and defensive state
&
\cellcolor{TableAltGray}
Attack-strategy pool and defensive memory
\\[1pt]

\bottomrule

\end{tabularx}}

\endgroup
\end{table*}

\subsection{Discussion}
\label{subsec:selfevolving_discussion}

Trustworthy self-evolution shifts robustness concerns from a single system state to the update process itself. Once accepted, an update can persist and shape later behavior or future modifications \cite{7_12,7_13}. Across the methods reviewed in this section, trustworthiness enters this process by restricting candidate changes, controlling which updates are retained, or adapting defenses as adversarial pressure evolves. The common concern is how an accepted change propagates through later versions of the system.
This perspective also changes the role of EC in trustworthy AI. In self-evolving systems, evolutionary state can persist across rounds through reusable populations or external stores \cite{harnessbank,7_10}. As a result, selection decisions made in one round affect the search that follows, linking present updates to the safety of future versions. Robustness therefore concerns the continuing update process as well as the quality of each evolved solution.

\section{Open Challenges and Future Research Directions}
\label{sec:future_directions}

\begin{figure}[!t]
    \centering
    \includegraphics[width=0.45\textwidth]{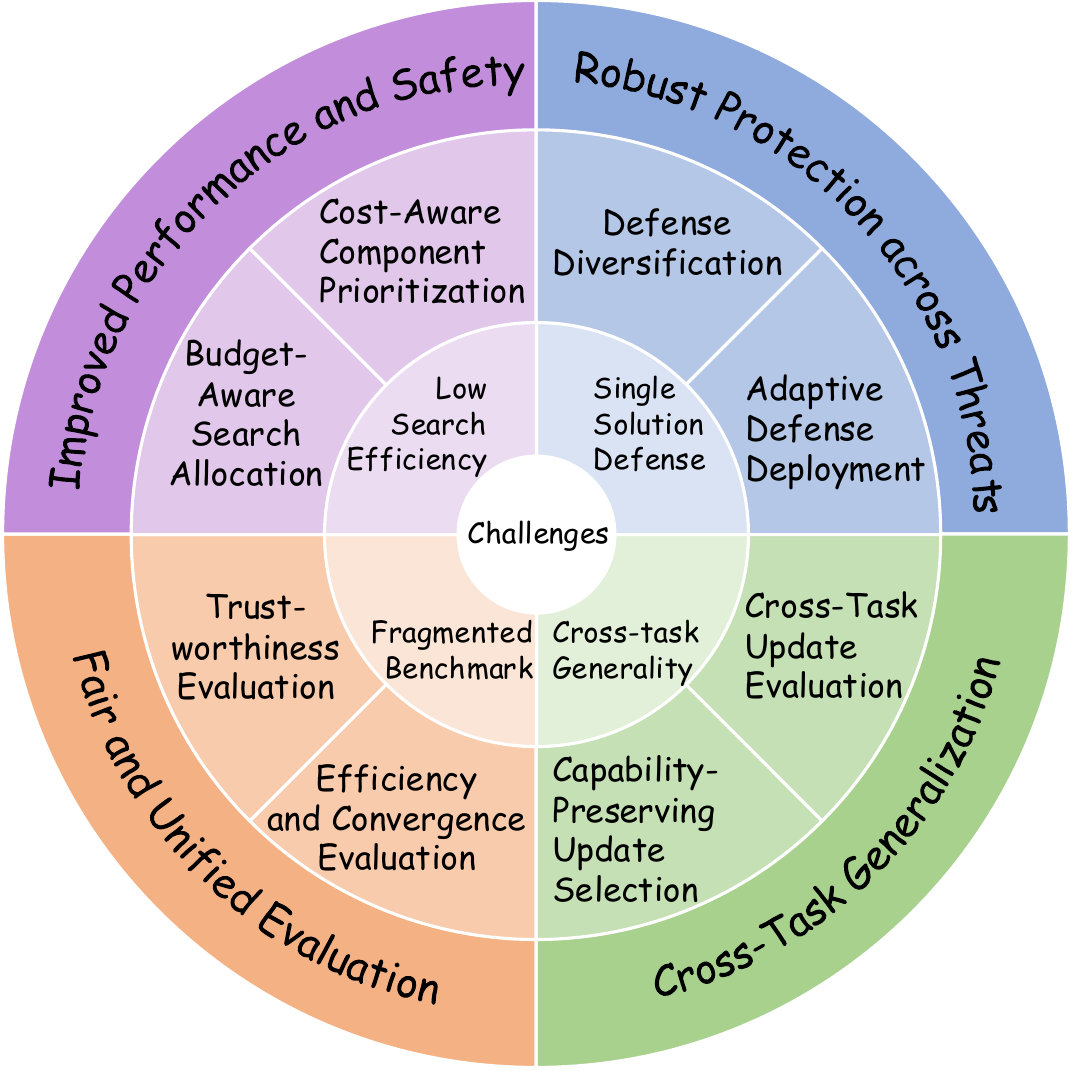}
    \caption{Roadmap linking open challenges to research directions and long-term goals for evolutionary computation in trustworthy AI.}
    \label{fig:future_research_roadmap}
\end{figure}

Based on the limitations identified in existing studies, Fig.~\ref{fig:future_research_roadmap} summarizes four major challenges in the use of EC for trustworthy AI and the corresponding future research directions. These challenges involve multi-component optimization, multi-solution defense, multi-task generalization, and multi-objective benchmarking. For each challenge, we outline possible directions that may improve efficiency, adaptability, robustness, and evaluation.

\subsection{Multi-Component Search under Limited Budgets}

Agentic models contain multiple components that can be optimized by EC, including prompts, retrieval modules, memory, tools, workflows, and model configurations. However, evolving an entire pipeline can be costly because EC requires repeated candidate evaluations and different components incur very different evaluation costs. Changes to prompts or routing rules may be tested with relatively few model queries, whereas updates to persistent memory or tools may require substantially more evaluation. Joint optimization therefore enlarges both the search space and the required budget.

Instead of optimizing all components jointly, EC could be applied only to selected parts of the pipeline. 
The cost of changing a component depends on the tokens and model queries needed to evaluate it and, for parameter updates, the computation required for fine-tuning. Future methods could prioritize components that are relatively inexpensive to evaluate but have a strong influence on performance or safety, and limit evolutionary search to these components. More expensive components could remain unchanged unless further improvement requires their modification. 
These challenges become more pronounced in agentic and multi-agent systems, where interactions among multiple agents can introduce coupled safety and security risks~\cite{romance,evo_marl}.

\subsection{Multi-Solution Defense for Diverse Threats}

Existing evolutionary defenses often search for a single final solution, such as one architecture, policy, training configuration, or safeguard. However, a model may face different data distributions, attack types, and safety requirements. A defense that is effective against one type of attack may perform poorly against another, making it difficult for a single solution to provide reliable protection across different settings.

A promising direction is to evolve multiple complementary solutions rather than selecting only one final defense. Different solutions can target different attacks, data conditions, or safety requirements, and can be combined through ensembles, voting, or context-dependent selection.

\subsection{Multi-Task Generalization in Self-Evolving AI}

Self-evolving agents often perform different tasks, including communication, retrieval, planning, coding, and tool use. However, evolutionary updates are often evaluated on only a limited set of tasks. An update that improves one task may not transfer to others and may even reduce performance on tasks that were previously handled well. 

Evolutionary updates can be evaluated across multiple tasks rather than only on the task that produced them. Updates that generalize across tasks while preserving previously learned capabilities are more suitable for long-term learning. This perspective is closely related to the stability--plasticity trade-off in continual learning~\cite{shi2025continual}.

\subsection{Multi-Objective Benchmarking for EC and Trustworthy AI}

Evaluating EC for trustworthy AI requires considering both trustworthiness and search efficiency. Improvements in robustness or safety may come with lower task performance or higher evaluation cost. However, existing benchmarks often assess trustworthy AI and evolutionary search separately, making it difficult to determine whether similar safety gains are achieved with comparable search effort.

A unified benchmark is needed to evaluate both task-level trustworthiness and search-level performance under the same evaluation budget. Comparisons can then consider robustness, safety, task performance, search cost, and convergence jointly rather than relying solely on a single final score. When useful, a weighted score may additionally summarize these dimensions, while the individual metrics and their trade-offs should also be reported.

\section{Conclusion}
\label{sec:conclusion}

This survey presents an overview of evolutionary computation for trustworthy AI, covering evolutionary attacks, evolutionary defenses, and trustworthy self-evolving AI. We first clarify the scope and positioning of this survey before introducing the key preliminaries and formulations. Next, we review evaluation and benchmarking in terms of both trustworthiness and evolutionary search performance. We then review how EC searches for failures across predictive, generative, and agentic systems and how it optimizes training data, model structures, learning strategies, and deployment safeguards for defense. Finally, we examine self-evolution from three perspectives: constraints on the evolution process, guidance for selecting which changes are preserved, and the co-evolution of attacks and defenses. Taken together, these sections organize the literature on EC for trustworthy AI around three recurring questions: what is evolved, how evolutionary search is performed, and how its outcomes are evaluated.

\bibliographystyle{IEEEtran}
\bibliography{egbib}

\end{document}